\documentclass{article} 
\usepackage{main,times}

\usepackage{amsmath,amsfonts,bm}

\def\eqref#1{equation~\ref{#1}}

\def\1{\bm{1}}

\DeclareMathAlphabet{\mathsfit}{\encodingdefault}{\sfdefault}{m}{sl}
\SetMathAlphabet{\mathsfit}{bold}{\encodingdefault}{\sfdefault}{bx}{n}

\usepackage{graphicx}
\definecolor{cvprblue}{rgb}{0.21,0.49,0.74}
\usepackage[breaklinks,colorlinks,citecolor=cvprblue]{hyperref}
\usepackage{url}
\usepackage{booktabs}
\usepackage[table]{xcolor}
\usepackage[dvipsnames]{xcolor}
\usepackage{wrapfig}
\usepackage{multirow}
\usepackage{algorithm, algpseudocode, amsmath, amssymb}

\definecolor{LightBlue}{RGB}{234,244,255}

\title{Mask Forcing: Improving Autoregressive Video Diffusion Distillation via Dual-Noise Masking Rollout}

\author{Zhuoran Zhao\textsuperscript{\normalfont1,2}, Shengju Qian\textsuperscript{\normalfont3}, Tongtong Liang\textsuperscript{\normalfont4}, Xianghao Kong\textsuperscript{\normalfont2}, Songchun Zhang\textsuperscript{\normalfont2}, \\
\bfseries Junchao Huang\textsuperscript{\normalfont5}, Guian Fang\textsuperscript{\normalfont6}, Xin Wang\textsuperscript{\normalfont3}, Pan Hui\textsuperscript{\normalfont1,2}, Anyi Rao\textsuperscript{\normalfont2} \\ \\
\textsuperscript{1}HKUST(GZ)
~\textsuperscript{2}HKUST
~\textsuperscript{3}LIGHTSPEED
~\textsuperscript{4}UCSD
~\textsuperscript{5}CUHK(SZ)
~\textsuperscript{6}NUS
}

\begin{document}

\maketitle

\begin{abstract}

Autoregressive (AR) video diffusion models have shown great potential in real-time video generation. Recent methods distill pretrained bidirectional video diffusion models into causal AR students through Distribution Matching Distillation (DMD), but the generated videos often suffer from over-saturation and over-smoothing issues, resulting in limited visual quality and realism. The key contributing factor is the mode-seeking behavior of the reverse KL objective in DMD, which can cause the student distribution to collapse onto only a few modes of the teacher distribution. To address this, we propose Mask Forcing, a Dual-Noise Masking Rollout strategy that perturbs the AR student self-rollout to mitigate mode collapse induced by reverse-KL mode seeking. The core idea is to inject cleaner signals into noisy rollout inputs via random masks along spatial and temporal axes during the self-rollout process of AR diffusion distillation. Such perturbations encourage the student rollouts to explore more regions of the teacher distribution, allowing DMD to provide learning signals beyond the modes already covered by the student. Moreover, the cleaner tokens act as denoising guidance for other noisier tokens, improving the intermediate rollout predictions and reducing error accumulation. Extensive experiments demonstrate that our method improves multiple AR video diffusion distillation methods with higher visual quality efficiently, without incorporating real video data or additional post-training stages. Project page: \href{https://alicezrzhao.github.io/mask_forcing/}{\textcolor{cvprblue}{https://alicezrzhao.github.io/mask-forcing}}.

\end{abstract}

\section{Introduction}

Video diffusion models have advanced rapidly in recent years, enabling the generation of long-duration and high-fidelity videos~\citep{wan2025, kong2024hunyuanvideo, hacohen2026ltx}. However, these models typically rely on bidirectional attention and multiple denoising timesteps, limiting their applicability to real-time streaming generation scenarios. 
Therefore, autoregressive (AR) diffusion models have emerged as a promising paradigm for real-time video generation~\citep{huang2026self, zhu2026causal, yang2025longlive, hu2026multiplayer, lingbot-world-v2}. By leveraging causal attention mechanisms, AR diffusion models can generate future chunks sequentially, conditioning each chunk on previously generated ones via causal dependencies. This formulation supports real-time inference and unbounded video generation without recomputing previously generated frames.

Teacher Forcing~\citep{ACDiT, gao2025ca2} and Diffusion Forcing~\citep{chen2024diffusion} are two representative paradigms for training causal autoregressive video diffusion models. Teacher Forcing trains the model to predict the next frames conditioned on the ground-truth frames, leading to exposure bias since the model can only condition on its own predictions during inference. Diffusion Forcing instead trains the model by assigning each frame with independently sampled noise. While this paradigm alleviates the distribution shift, it still fails to align training with inference. Self Forcing~\citep{huang2026self} bridges the train-test gap by training the model with self-rollout, generating the next frame based on previously self-generated frames rather than ground-truth context and distilling the teacher's knowledge via a DMD loss~\citep{yin2024one}. However, a critical limitation of combining self-rollout with a DMD loss is that it often produces over-saturated and over-smoothed videos, exhibiting low visual quality and limited realism.

This phenomenon can be attributed to two key factors. First, the reverse-KL objective in DMD exhibits mode-seeking behavior that tends to cover only the high-probability regions of the teacher's distribution~\citep{chen2025retaining, cai2026mode, zheng2026large}. This behavior induces mode collapse, leading to reduced diversity in the generated videos. Second, the intermediate rollout predictions receive no explicit training signal at each step and suffer from error accumulation. Recent methods mitigate these issues by explicitly incorporating real data into the training objective~\citep{yin2024improved, chen2026data, liu2026opsd}, but they rely on complex data-curation processes or additional post-training, which further complicates the multi-stage training pipeline. Another line of research incorporates reinforcement learning post-training to improve the quality of the distilled AR model~\citep{zhang2026astrolabe}, but its performance still relies on reward models and is sensitive to multiple hyperparameters. A complementary line of work balances mode-seeking and mode-covering objectives~\citep{cai2026mode, zheng2026large, li2026distillalign}, but the generated videos can still appear over-saturated and lack fine-grained visual details.
This begs the question: can we improve AR video diffusion distillation without real data curation or additional post-training stages?

\begin{figure}
    \centering
    \includegraphics[width=1\linewidth]{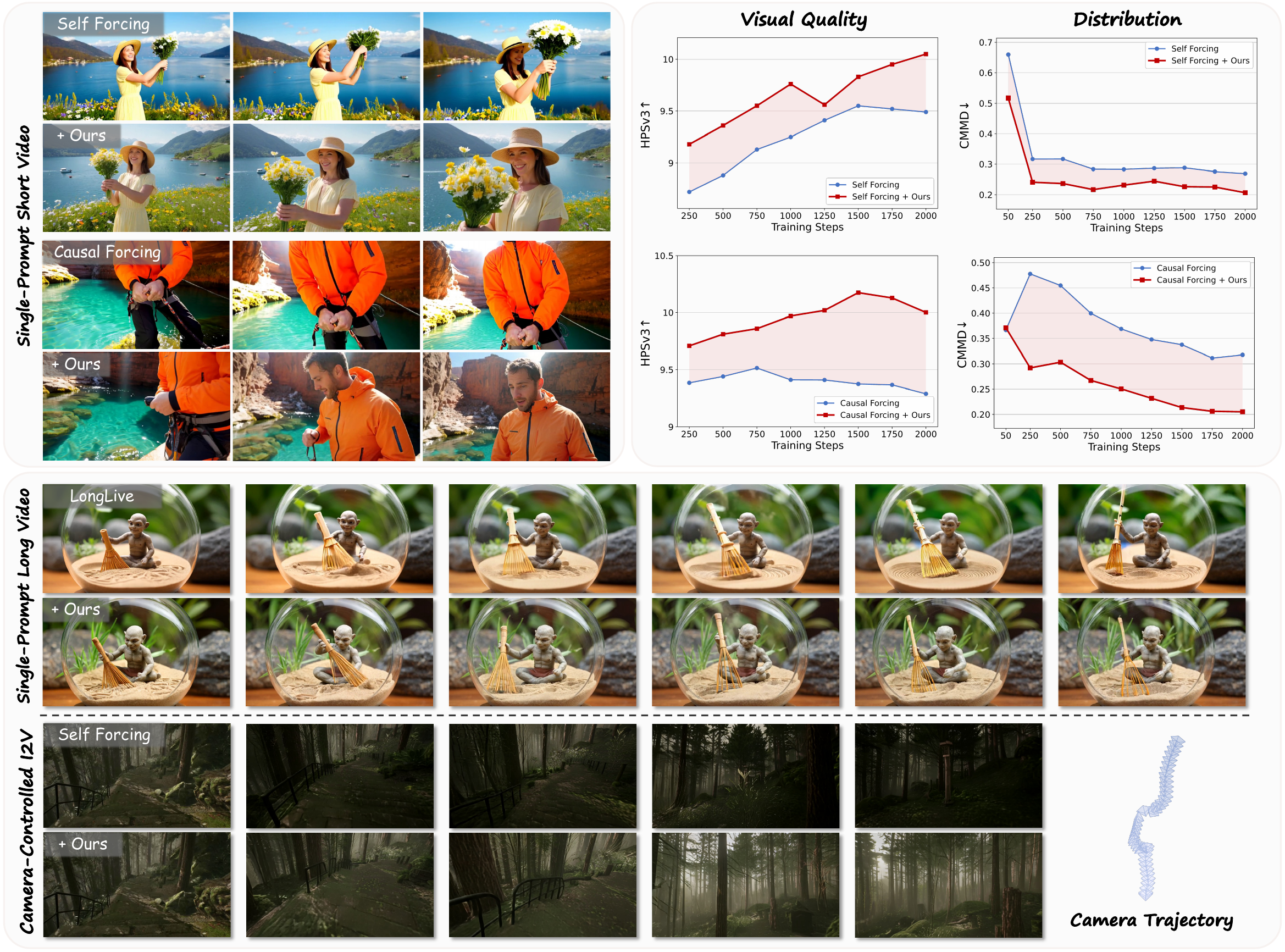}
    \vspace{-10pt}
    \caption{Mask Forcing improves autoregressive video generation through a Dual-Noise Masking Rollout strategy that alleviates the mode-seeking behavior of the reverse KL objective in DMD, enhancing video visual quality efficiently without incorporating real video data or additional post-training. We show consistent improvement in visual realism across short-video, long-video, and camera-controlled autoregressive generation.}
    \label{fig:teaser}
\end{figure}

To address this question, we focus on the self-rollout trajectory, which determines both the student samples exposed to DMD and the intermediate predictions reused in subsequent denoising and autoregressive conditioning.
Perturbing the rollout trajectory can expose DMD to a broader range of student samples, allowing it to provide learning signals beyond the modes the student already covers. Moreover, cleaner tokens can serve as context for denoising noisier tokens~\citep{chefer2026self}, helping improve intermediate rollout predictions and reduce error accumulation. Based on these motivations, we propose Mask Forcing, which injects randomly masked cleaner tokens into the rollout input to mitigate both mode-seeking behavior and error accumulation. At each rollout transition, Mask Forcing randomly samples a mask and an additional timestep corresponding to a lower noise level than the original one and injects the cleaner tokens into the rollout input. Specifically, the latent positions selected by the mask are re-noised to this lower noise level, while the remaining positions are kept at the original level, producing a dual-noise rollout input. The conditioning timestep fed to the model remains the original one, requiring the model to denoise inputs whose local noise levels are partially inconsistent with the global timestep. Randomizing both the mask and the lower-noise timestep diversifies the student rollout trajectory and broadens the region of sample space explored during training, enabling DMD to provide learning signals from teacher modes not yet covered by the student. Beyond perturbing the rollout trajectory, Mask Forcing uses lower-noise tokens as context for denoising noisier tokens. This design is consistent with the observation in Self-Flow~\citep{chefer2026self} that cleaner context in mixed-noise inputs can facilitate denoising, thereby improving intermediate predictions and reducing error accumulation during self-rollout.

Extensive experiments across multiple AR video distillation methods validate the effectiveness of our method in both chunk-wise and frame-wise settings. Representative results in Fig.~\ref{fig:teaser} show that our method significantly improves the visual quality against various baselines, with enhanced realism and richer high-frequency details. Comprehensive evaluations demonstrate improvements across multiple benchmarks. In summary, our contributions are as follows:
\begin{itemize}
    \item We propose Mask Forcing, a simple and effective approach that alleviates the mode-seeking behavior of the reverse-KL objective in self-rollout DMD training for AR video diffusion distillation, without incorporating real video data or additional post-training stages.
    \item We introduce a Dual-Noise Masking Rollout strategy that injects lower-noise signals into noisy rollout inputs through random masks along spatial and temporal axes. Such perturbations diversify student rollout trajectories to cover more teacher modes, while providing cleaner context for denoising to reduce error accumulation.
    \item Extensive experiments on multiple AR video distillation methods demonstrate the effectiveness of Mask Forcing in both chunk-wise and frame-wise settings, with significantly improved visual quality and faster convergence. Comprehensive ablations further validate the effects of different masking mechanisms.
\end{itemize}

\section{Related Work}

\subsection{Autoregressive Video Generation}

Autoregressive video diffusion models enable real-time video generation by sequentially generating videos conditioned on historical context. Teacher Forcing~\citep{ACDiT, gao2025ca2} denoises the current chunk conditioned on clean ground-truth context, which suffers from a train-test gap and exposure bias. Diffusion Forcing~\citep{chen2024diffusion} assigns each frame with independently sampled noise to approximate rollout distributions, but still fails to align training with inference. Self Forcing~\citep{huang2026self} bridges this gap by performing AR self-rollout on self-generated histories and distills a bidirectional teacher into the causal student via a DMD loss. Causal Forcing~\citep{zhu2026causal} further uses an AR teacher for ODE initialization to reduce the architecture gap. LongLive~\citep{yang2025longlive} extends causal AR generation to long videos via short window attention with frame sink and streaming long tuning. Despite these advances, distilled AR models still exhibit limited visual quality and realism, motivating methods that introduce additional training signals, real data, or post-training stages. DMD2~\citep{yin2024improved} introduces a GAN loss and real training data, but suffers from training instability and additional real data curation. DFD~\citep{chen2026data} integrates real data into the distillation score, but requires post-training on a DMD2-pretrained model. Astrolabe~\citep{zhang2026astrolabe} explores RL post-training on distilled AR models, but performance is limited by reward models. In contrast, Mask Forcing mitigates mode collapse by perturbing student rollouts, improving AR distillation without real video supervision or post-training.

\subsection{Mode Seeking and Mode Covering in Video Diffusion Distillation}

DMD~\citep{yin2024one} and DMD2~\citep{yin2024improved} use score-based reverse KL distribution matching to align student-generated samples with the teacher distribution. This mode-seeking objective can improve sample fidelity but may concentrate the student distribution on a limited subset of teacher modes. DMD2 further introduces an adversarial loss on real data. In contrast, trajectory-based consistency objectives are commonly associated with the mode-covering behavior of forward divergence~\citep{song2023consistency,kim2024consistency,lu2025simplifying}. Such objectives can cover more teacher modes, but may average across modes and exhibit lower sample quality. Recent methods balance these behaviors by combining complementary objectives. Mode Seeking meets Mean Seeking~\citep{cai2026mode} uses separate heads for supervised flow matching on long videos and reverse-KL distribution matching against a short-video teacher. rCM~\citep{zheng2026large} augments continuous-time consistency with score distillation regularization. DistillAlign~\citep{li2026distillalign} analyzes the initialization effects and jointly optimizes DMD and a consistency distillation loss. Unlike these methods, Mask Forcing retains the original DMD objective and instead perturbs the student rollouts via dual-noise masking. This broadens the student distribution exposed to DMD, allowing it to provide learning signals from teacher modes not reached by standard self-rollout.

\subsection{Masked Modeling}

Masked modeling has become a powerful paradigm for representation and generative learning in computer vision~\citep{he2022masked, bao2021beit, chang2022maskgit}. The core idea is to mask a portion of the input and train the model to recover it. In representation learning, MAE~\citep{he2022masked} masks a large portion of random patches from the input image and reconstructs them in pixel space by leveraging context from visible parts. In generative modeling, MaskGIT~\citep{chang2022maskgit} adopts a mask-then-predict objective with parallel iterative decoding to synthesize images. Masking can also be realized through heterogeneous noise levels that control the information retained by each token. For multi-modal generation, Self-Flow~\citep{chefer2026self} introduces a self-supervised framework for flow matching that combines mix-timestep scheduling with masking for representation alignment. In AR generation, Diffusion Forcing associates each frame with a random, independent noise level, which can be regarded as partial masking along the time axis. Inspired by these works, we introduce dual-noise masking into AR video distillation to diversify self-rollout trajectories and provide cleaner context for denoising noisier tokens.

\section{Method}

\subsection{Preliminaries}

\noindent \textbf{Autoregressive (AR) Video Generation.}
An AR video model represents a video as a sequence of $F$ chunks $x_{1:F}=(x_1,\ldots,x_F)$, where each chunk may include one or more latent frames. It factorizes the text-conditioned joint distribution as $p(x_{1:F}\mid c)=\prod_{i=1}^{F}p(x_i\mid x_{<i},c)$, where $c$ denotes the text prompt and $x_{<i}=(x_1,\ldots,x_{i-1})$ denotes the preceding video chunks. Each conditional distribution $p(x_i\mid x_{<i},c)$ is modeled using a diffusion process following the flow-matching formulation, where the noisy sample is defined as $x_i^t=(1-t)x_i+t\epsilon_i$, with $\epsilon_i\sim\mathcal{N}(0,I)$ and $t\in[0,1]$. Each video chunk is generated through this denoising process conditioned on $c$ and the historical context $x_{<i}$, which is stored in the key-value (KV) cache. Teacher Forcing (TF) and Diffusion Forcing (DF) are two typical training paradigms for AR video models using frame-wise MSE loss between predicted and ground-truth targets. In TF, the timestep $t$ is shared across all frames, and the context consists of clean ground-truth frames. In DF, each frame is assigned an independently sampled timestep $t_i$, and the historical context is noisy. To mitigate the train-test gap and alleviate error accumulation, Self Forcing unrolls the model on its own generated samples during training, with $x_{1:F}^{\theta}\sim\prod_{i=1}^{F}p_\theta(x_i\mid x_{<i},c)$.

\begin{figure}
    \centering
    \includegraphics[width=1\linewidth]{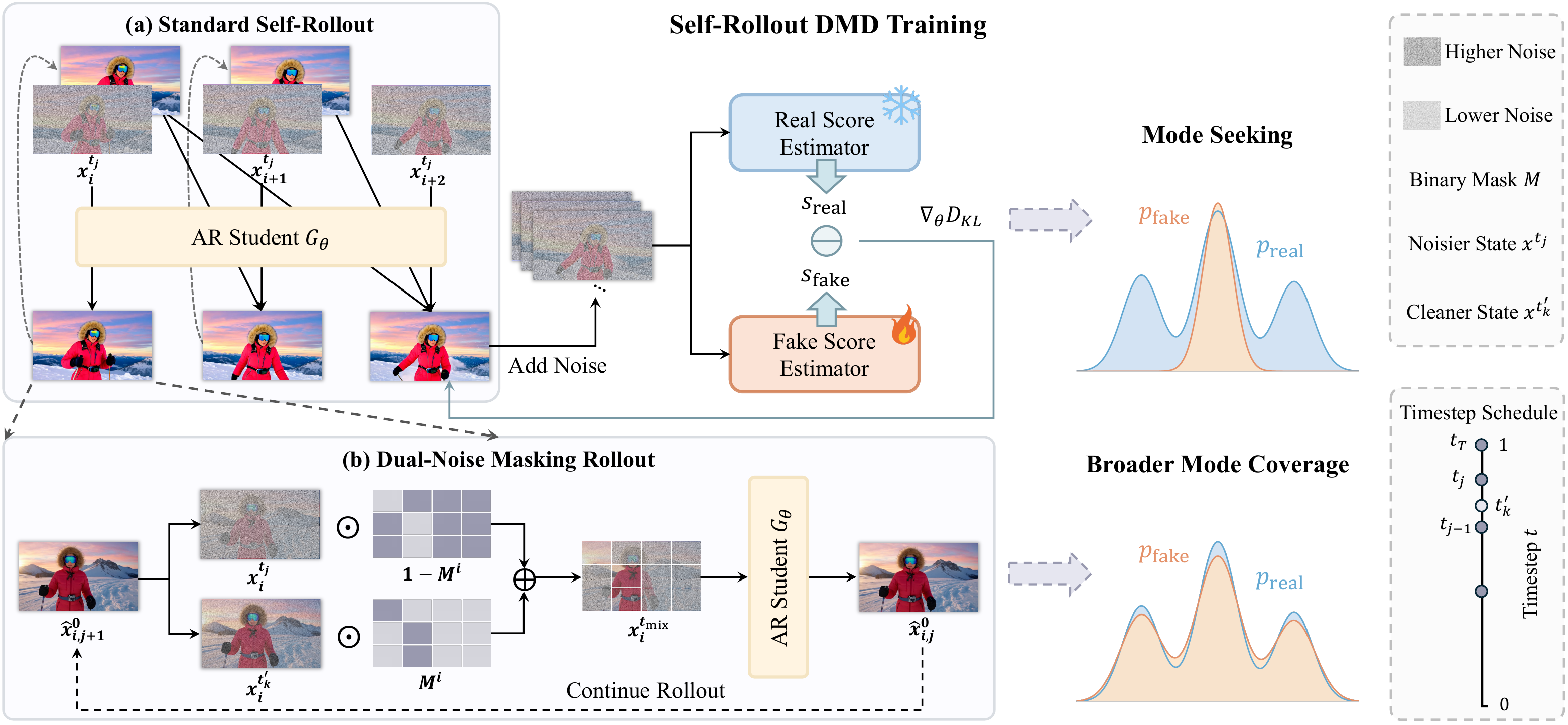}
    \vspace{-10pt}
    \caption{(a) Standard self-rollout DMD training suffers from mode collapse due to the mode-seeking reverse-KL objective, resulting in over-saturation and limited realism. (b) Mask Forcing perturbs the student rollout trajectory through dual-noise masking rollout, encouraging broader teacher-mode coverage while providing cleaner context for denoising noisy tokens.}
    \label{fig:forcing_comp}
\end{figure}

\noindent \textbf{Distribution Matching Distillation (DMD).} DMD~\citep{yin2024one, yin2024improved} distills a pretrained multi-step teacher model into a few-step student model $G_{\theta}$ by minimizing the reverse KL divergence from the student generator induced distribution $p_\mathrm{fake}$ to the teacher distribution $p_\mathrm{real}$. Specifically, DMD adopts a reverse KL objective, whose gradient is used to update the student model:
\begin{equation}
\nabla_{\theta}{\mathcal{L}_\mathrm{DMD}} = \mathbb{E}{[\nabla_{\theta}D_{KL}( p_{\mathrm{fake}, t} || p_{\mathrm{real}, t})]},
\end{equation}
\begin{equation}
D_{KL}( p_{\mathrm{fake}, t} || p_{\mathrm{real}, t}) = \mathbb{E}{\left[\log\frac{p_{\mathrm{fake}, t}(x_t)}{p_{\mathrm{real}, t}(x_t)}\right] = -\mathbb{E}[\log p_{\mathrm{real}, t}(x_t) - \log p_{\mathrm{fake}, t}(x_t)]},
\end{equation}
where $x_t$ is a noisy sample corresponding to timestep $t$: $x_t = (1-t)G_{\theta}(z) + t \epsilon$ with $\epsilon \sim \mathcal{N}(0, I)$, and $z$ is latent drawn from random noise. The gradient to update $G$ is formulated as the difference between two score functions:
\begin{equation}
\nabla_{\theta}{\mathcal{L}_\mathrm{DMD}} = \mathbb{E}\left[ - (s_\mathrm{real}(x_t, t) - s_\mathrm{fake}(x_t, t))\frac{dG}{d\theta} \right],
\end{equation}
where $s_\mathrm{real}(x_t, t) = \nabla_{x_t}\log p_{\mathrm{real}, t}(x_t)$, $s_\mathrm{fake}(x_t, t) = \nabla_{x_t}\log p_{\mathrm{fake}, t}(x_t)$ denote the scores of the teacher and student distributions, respectively. However, the reverse KL objective is inherently mode-seeking and may concentrate the student distribution on a limited set of teacher modes, leading to over-saturation and reduced realism~\citep{chen2026data}.

\subsection{Dual-Noise Masking Rollout}

\subsubsection{Overview}

We propose a dual-noise masking rollout strategy for the self-rollout DMD training to mitigate mode collapse induced by the mode-seeking behavior of reverse KL in AR diffusion distillation. The core idea is to inject low-noise signals into noisy rollout inputs during the rollout process via random masks applied both within and across chunks. These cleaner signals serve two purposes. First, they perturb the student rollouts, encouraging broader coverage of high-density regions in the teacher distribution and thereby mitigating mode collapse and visual artifacts such as over-saturation and over-smoothing. Second, cleaner tokens provide context for denoising noisier tokens, following the principle of masked modeling. An overview of our method is shown in Fig.~\ref{fig:forcing_comp}.

\subsubsection{Self-Rollout with Dual-Noise Masking}

An AR video diffusion model represents a video as a sequence of $F$ chunks $x = (x_1, x_2, ..., x_F)$, where each chunk contains one or more latent frames. Using the flow-matching formulation, a noisy chunk is defined as:
\begin{equation}
x_i^{t}=(1-t)\,x_i^{0}+t\,\epsilon,\qquad \epsilon\sim\mathcal{N}(0,I),
\end{equation} 
where $t\in[0,1]$ denotes the timestep, which interpolates between the clean chunk at $t=0$ and pure Gaussian noise at $t=1$. During the self-rollout training, the student model $G_{\theta}$ generates the video chunk by chunk and each chunk is produced by iterative denoising over a fixed schedule of $T$ timesteps $\{t_1,\ldots,t_T\}$ selected from the $N_t = 1000$ training timesteps, where $t_T=1$ denotes pure noise, $t_0=0$ denotes the clean output. We use $T=4$ denoising steps throughout training. For each chunk index $i$, at denoising timestep $t_j$, the student model predicts a clean chunk estimate $\hat{x}_{i,j}^0$ from the noisy chunk $x_i^{t_j}$, conditioned on the previously generated clean chunks $x_{<i}$, the text prompt $c$, and the timestep $t_j$:
\begin{equation}
\hat{x}_{i,j}^0 = G_\theta\left(x_i^{t_j} \mid x_{<i}, c, t_j\right).
\end{equation}
The DMD objective is evaluated only on the completed self-rollout, providing no explicit training signal for intermediate predictions at each step. Since intermediate predictions are reused in subsequent denoising steps and as context for later chunks, their errors can accumulate throughout the rollout. Meanwhile, the reverse-KL objective is inherently mode-seeking and concentrates the student rollout distribution on a narrow set of high-density modes of the teacher distribution. Together, these limitations may contribute to degraded visual quality and realism in generated videos.

We address this issue by injecting cleaner signals into the student model input to perturb the student rollout trajectory, encouraging the student rollouts to reach more high-density regions of the teacher distribution. Specifically, we adopt a dual-timestep scheduling strategy inspired by Self-Flow~\citep{chefer2026self}. We retain the denoising schedule of the base model and define $\Delta$ as the size of the timestep window measured in training timestep units, corresponding to a normalized width of $\Delta/N_t$. At each denoising timestep $t_j$, we uniformly sample an additional timestep $t'_k$ within the corresponding window in timestep space:
\begin{equation}
\ell_j = \max\left(t_{\min},\;t_j-\frac{\Delta}{N_t}\right),
\end{equation}
\begin{equation}
t'_k \sim \mathrm{Uniform}\big([\ell_j,t_j]\big).
\end{equation}
Here, $t'_k$ corresponds to a lower noise level than $t_j$, and the floor $t_{\min}$ prevents the injected signal from being nearly clean.

For each chunk $i$, we construct a binary mask $M^i$ with a masking ratio $\alpha$. The mask is sampled independently across frames within a chunk (spatial axis) and across chunks (temporal axis), so that different frames of the same chunk and different chunks receive different dual-noise patterns. We observe that such mask diversity affects the visual quality and motion dynamics balance of the generated videos. At the $j$-th denoising step, corresponding to timestep $t_j$, we sample $\epsilon_j\sim\mathcal{N}(0,I)$ and use it to re-noise the clean prediction from the previous step $\hat{x}_{i,j+1}^{0}$ at $t'_k$ and $t_j$, yielding the cleaner sample $x_i^{t'_k}$ and the noisier sample $x_i^{t_j}$, respectively:
\begin{equation} 
x_i^{t'_k} = (1-t'_k)\,\hat{x}_{i,j+1}^{0} + t'_k\,\epsilon_j, 
\end{equation} 
\begin{equation} 
x_i^{t_j} = (1-t_j)\,\hat{x}_{i,j+1}^{0} + t_j\,\epsilon_j. 
\end{equation}
We use the random mask $M^i$ to select tokens from the cleaner sample at masked positions ($M^i=1$), while retaining tokens from the original noisier sample at the remaining positions. This yields the dual-noise input:
\begin{equation}
x_i^{t_\mathrm{mix}} = M^i \odot x_i^{t'_k} + (1 - M^i) \odot x_i^{t_j}.
\end{equation}
Consequently, the student generator $G_\theta$ receives the dual-noise input at denoising step $j$ and produces the updated clean prediction:
\begin{equation}
\hat{x}_{i,j}^{0} = G_\theta\left(x_i^{t_\mathrm{mix}} \mid x_{<i}, c, t_j\right).
\end{equation}
The student generator is conditioned on the scheduled timestep $t_j$, while cleaner tokens in the dual-noise input are treated as a training perturbation. The model follows the original denoising schedule at inference. At the next denoising step, $\hat{x}_{i,j}^{0}$ is re-noised at the scheduled timestep $t_{j-1}$ and a newly sampled cleaner timestep within its corresponding timestep window. The two noisy samples are then combined using the mask $M^i$ to form the next dual-noise input, replacing the standard re-noising operation in self-rollout. Once the current chunk reaches the sampled exit step $s$, the corresponding clean prediction $\hat{x}_{i,s}^{0}$ is used to update the KV cache and condition subsequent chunks.

This dual-noise perturbation achieves two goals. First, it diversifies the student rollout trajectories, so that the distillation score in the DMD loss is evaluated over a broader region of the sample space rather than a narrow subset of teacher modes. This encourages the student model to explore more diverse modes and prevents it from collapsing onto the high-density modes of the teacher model induced by the reverse KL objective. Second, since the masked tokens carry lower noise, the student model is guided to exploit them as context, helping denoise the noisier tokens and improving intermediate rollout predictions.

\subsubsection{Distributional Analysis}

We analyze the student rollout distribution induced by dual-noise masking. We characterize it as a mixture over masking trajectories and decompose its reverse-KL objective using mutual information.

Fix a text condition $c$ and a DMD score-noising timestep $\tau$, distinct from the denoising timesteps used during self-rollout. Let $X_\tau$ denote the completed student rollout noised at $\tau$ and let $p_\tau$ denote the corresponding teacher noisy marginal. The masking trajectory $V$ is defined as the collection of all masks and lower-noise timesteps sampled during the rollout, with $V\sim\pi$, where $\pi$ is determined by the mask ratio, timestep window, and mask sampling scheme.

Conditioned on a masking trajectory $V=v$, $q_{\theta,\tau}^{v}(x)$ denotes the corresponding student distribution.
Marginalizing over $V$ gives the masked-rollout distribution $\bar q_{\theta,\tau}(x):=\mathbb{E}_{V\sim\pi}[q_{\theta,\tau}^{V}(x)]$. 
The dependence between $V$ and $X_\tau$ conditioned on $c$ is measured by their mutual information, given by
\begin{equation}
I_\theta(V;X_\tau\mid c)=\mathbb{E}_{V\sim\pi}\left[D_{\mathrm{KL}}\left(q_{\theta,\tau}^{V}\|\bar q_{\theta,\tau}\right)\right].
\end{equation}
The reverse-KL objective of the masked-rollout distribution can be decomposed as
\begin{equation}
D_{\mathrm{KL}}\left(\bar q_{\theta,\tau}\|p_\tau\right)
=
\mathbb{E}_{V\sim\pi}
\bigl[
D_{\mathrm{KL}}\left(q_{\theta,\tau}^{V}\|p_\tau\right)
\bigr]
-
I_\theta(V;X_\tau\mid c).
\end{equation}
$D_{\mathrm{KL}}(\bar q_{\theta,\tau}\|p_\tau)$ is the divergence between the marginal masked-rollout distribution and the teacher marginal. $\mathbb{E}_{V\sim\pi}[D_{\mathrm{KL}}(q_{\theta,\tau}^{V}\|p_\tau)]$ is the average divergence between each trajectory-conditioned distribution and the teacher marginal, while $I_\theta(V;X_\tau\mid c)$ measures their output diversity.

When different masking trajectories induce distinct conditional distributions, $I_\theta(V;X_\tau\mid c)>0$. Individual trajectory-conditioned distributions may remain locally concentrated on different teacher-supported regions, while their marginal mixture can collectively cover these regions. This explains how marginalizing over trajectories can alleviate mode-seeking behavior in the overall student distribution without requiring individual conditional distributions to be mode-covering. The detailed derivation is provided in Appendix~\ref{app:theory}.

The mask ratio, timestep window, and mask sampling scheme determine $\pi$ and the strength of the induced rollout perturbations. Under weak perturbations, different masking trajectories induce similar output distributions, resulting in small $I_\theta(V;X_\tau\mid c)$, and limited additional teacher-mode coverage. Stronger perturbations can produce more distinct output distributions and increase $I_\theta(V;X_\tau\mid c)$, but excessive perturbations can move them away from teacher-supported regions and increase their average reverse KL. These factors balance trajectory diversity and teacher alignment. We study their effects through ablations in Sec.~\ref{sec:ablation}.

\section{Experiments}

\noindent \textbf{Implementation Details.}
We adopt Self Forcing~\citep{huang2026self}, LongLive~\citep{yang2025longlive}, and Causal Forcing~\citep{zhu2026causal} as our baselines, using Wan2.1-T2V-1.3B~\citep{wan2025} as the base model and Wan2.1-T2V-14B as the teacher. Our method is applied during the self-rollout DMD training stage. The generated videos consist of 81 frames at a resolution of 832 $\times$ 480, and the training prompts are sampled from the VidProM dataset~\citep{wang2024vidprom}. We implement each baseline under both chunk-wise and frame-wise autoregressive generation settings, with 3 latent frames per chunk in the chunk-wise setting. Since Self Forcing and LongLive do not release their frame-wise ODE initialization models, we train these models using ODE-paired data distilled from the bidirectional teacher. For Causal Forcing, we use its causal student model initialized via ODE distillation from the autoregressive teacher. During training, we set the mask ratio $\alpha$ to 0.2, the timestep window $\Delta$ to 250, and the floor $t_{\min}$ to the value at schedule index 20. In the chunk-wise setting, the mask is sampled independently across frames within a chunk and across chunks, while in the frame-wise setting, the mask is sampled independently across chunks. This training procedure takes approximately 1.5k steps and 14 hours on 8 GPUs.

\begin{figure}
    \centering
    \includegraphics[width=1\linewidth]{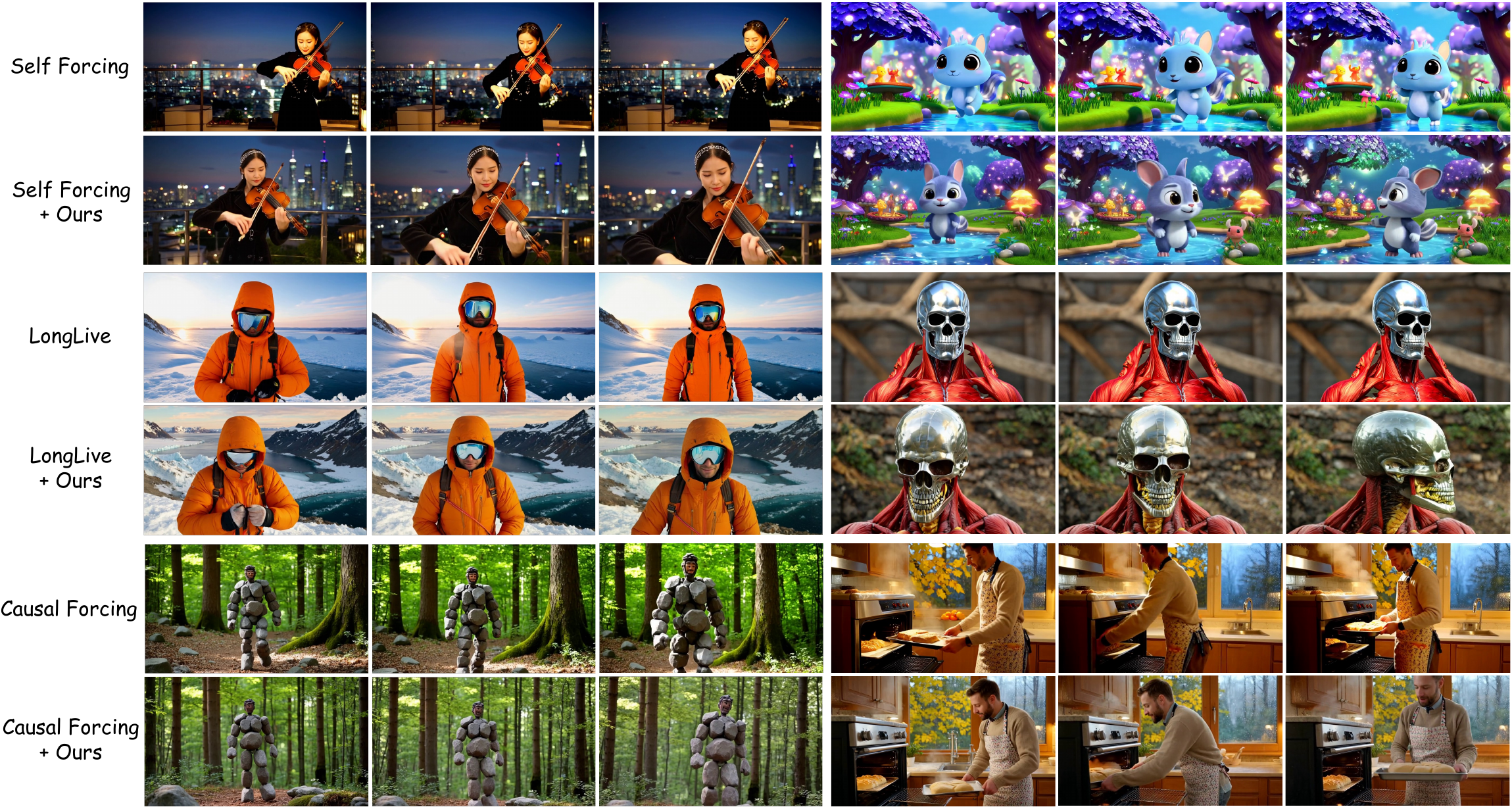}
    \caption{Qualitative comparison of our method (+Ours) against different baselines. Visual results show that our method generates videos with higher visual quality and greater realism, exhibiting fewer over-saturation artifacts and richer high-frequency details.}
    \label{fig:quali_1}
\end{figure}

\begin{table}[t]
\centering
\caption{Quantitative results on the 100-prompt set and VBench benchmarks under chunk-wise and frame-wise settings.}
\label{tab:main}
\resizebox{\columnwidth}{!}{%
\begin{tabular}{lccccc|ccc}
\toprule
Method & HPSv3 $\uparrow$ & Vision. $\uparrow$ & Instruct. $\uparrow$ & MQ $\uparrow$ & Dynamic. $\uparrow$ & Total $\uparrow$ & Quality $\uparrow$ & Semantic $\uparrow$  \\
\midrule
\rowcolor{gray!15}
\multicolumn{9}{l}{\textit{Chunk-wise}} \\
Self Forcing & 9.55 & 10.10 & 38.50 & 15.88 & 70 & 81.89 & 82.99 & 77.49 \\
\rowcolor{LightBlue}
\hspace{1em}+ Ours & 9.84$_{\textcolor{ForestGreen}{+.29}}$ & 11.37$_{\textcolor{ForestGreen}{+1.27}}$ & 45.03$_{\textcolor{ForestGreen}{+6.53}}$ & 20.49$_{\textcolor{ForestGreen}{+4.61}}$ & 82$_{\textcolor{ForestGreen}{+12}}$ & 82.61$_{\textcolor{ForestGreen}{+.72}}$ & 83.68$_{\textcolor{ForestGreen}{+.69}}$ & 78.74$_{\textcolor{ForestGreen}{+1.25}}$ \\
Causal Forcing & 9.37 & 10.36 & 40.41 & 17.73 & 76 & 82.67 & 83.58 & 78.98 \\
\rowcolor{LightBlue}
\hspace{1em}+ Ours & 10.17$_{\textcolor{ForestGreen}{+.80}}$ & 11.58$_{\textcolor{ForestGreen}{+1.22}}$ & 46.30$_{\textcolor{ForestGreen}{+5.89}}$ & 21.54$_{\textcolor{ForestGreen}{+3.81}}$ & 82$_{\textcolor{ForestGreen}{+6}}$ & 82.76$_{\textcolor{ForestGreen}{+.09}}$ & 83.69$_{\textcolor{ForestGreen}{+.11}}$ & 79.01$_{\textcolor{ForestGreen}{+.03}}$ \\
LongLive & 9.11 & 10.77 & 42.48 & 21.20 & 76 & 82.02 & 82.87 & 78.66 \\
\rowcolor{LightBlue}
\hspace{1em}+ Ours & 10.14$_{\textcolor{ForestGreen}{+1.03}}$ & 11.00$_{\textcolor{ForestGreen}{+.23}}$ & 42.65$_{\textcolor{ForestGreen}{+.17}}$ & 22.34$_{\textcolor{ForestGreen}{+1.14}}$ & 69$_{\textcolor{red}{-7}}$ & 82.75$_{\textcolor{ForestGreen}{+.73}}$ & 83.71$_{\textcolor{ForestGreen}{+.84}}$ & 78.91$_{\textcolor{ForestGreen}{+.25}}$ \\
\bottomrule
\rowcolor{gray!15}
\multicolumn{9}{l}{\textit{Frame-wise}} \\
Self Forcing & 9.34 & 9.45 & 35.62 & 18.27 & 53 & 80.73 & 81.72 & 76.78 \\
\rowcolor{LightBlue}
\hspace{1em}+ Ours & 9.79$_{\textcolor{ForestGreen}{+.45}}$ & 10.49$_{\textcolor{ForestGreen}{+1.04}}$ & 39.60$_{\textcolor{ForestGreen}{+3.98}}$ & 19.07$_{\textcolor{ForestGreen}{+.80}}$ & 61$_{\textcolor{ForestGreen}{+8}}$ & 81.49$_{\textcolor{ForestGreen}{+.76}}$ & 82.55$_{\textcolor{ForestGreen}{+.83}}$ & 77.23$_{\textcolor{ForestGreen}{+.45}}$ \\
Causal Forcing & 9.67 & 10.58 & 37.56 & 20.25 & 28 & 80.64 & 81.55 & 77.00 \\
\rowcolor{LightBlue}
\hspace{1em}+ Ours & 9.96$_{\textcolor{ForestGreen}{+.29}}$ & 10.71$_{\textcolor{ForestGreen}{+.13}}$ & 39.60$_{\textcolor{ForestGreen}{+2.04}}$ & 23.69$_{\textcolor{ForestGreen}{+3.44}}$ & 52$_{\textcolor{ForestGreen}{+24}}$ & 82.28$_{\textcolor{ForestGreen}{+1.64}}$ & 83.19$_{\textcolor{ForestGreen}{+1.64}}$ & 78.62$_{\textcolor{ForestGreen}{+1.62}}$ \\
LongLive & 9.19 & 9.35 & 38.50 & 12.14 & 25 & 80.97 & 81.86 & 77.41 \\
\rowcolor{LightBlue}
\hspace{1em}+ Ours & 9.46$_{\textcolor{ForestGreen}{+.27}}$ & 10.55$_{\textcolor{ForestGreen}{+1.20}}$ & 42.78$_{\textcolor{ForestGreen}{+4.28}}$ & 19.32$_{\textcolor{ForestGreen}{+7.18}}$ & 76$_{\textcolor{ForestGreen}{+51}}$ & 81.47$_{\textcolor{ForestGreen}{+.50}}$ & 82.33$_{\textcolor{ForestGreen}{+.47}}$ & 78.05$_{\textcolor{ForestGreen}{+.64}}$ \\
\bottomrule
\end{tabular}%
}
\end{table}

\noindent \textbf{Evaluation.}
We adopt the 100-prompt set with rich motion and complex actions from Causal Forcing and VBench as our primary evaluation benchmarks. For the 100-prompt set, we use HPSv3~\citep{ma2025hpsv3} to evaluate the overall visual quality and employ VisionReward~\citep{xu2026visionreward} to report vision reward score (Vision.), instruction following (Instruct.), and motion quality (MQ) sub-scores following Causal Forcing. Dynamic Degree (Dynamic.) is used to evaluate motions in video with RAFT~\citep{teed2020raft}. All metrics are scaled by 100 for readability except for HPSv3. For VBench, we report the Total Score, Quality Score, and Semantic Score with the 946 standard prompts. For the long-video generation setting, we additionally evaluate our method against LongLive with the first 100 prompts from the MovieGen~\citep{polyak2024movie} extended version and the VBench-Long official prompt set on 30-second video generation following LongLive.

\begin{figure}[t]
    \centering
    \includegraphics[width=1\linewidth]{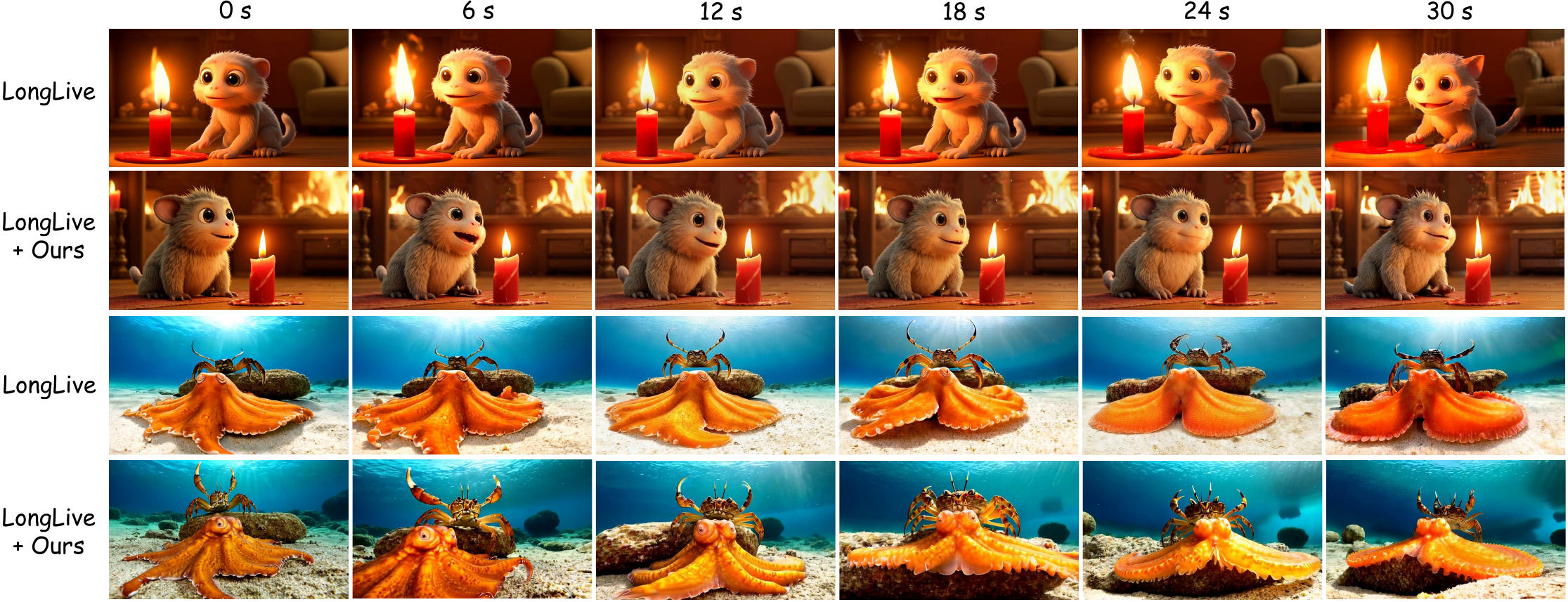}
    \caption{Qualitative results of single-prompt long-video setting on LongLive. Incorporating our method (+Ours) enhances the video generation quality with more visual details.}
    \label{fig:quali_long}
\end{figure}

\begin{table}[t]
\centering
\caption{Single-prompt 30s long video generation on MovieGen and VBench-Long benchmarks.}
\label{tab:long}
\resizebox{1\linewidth}{!}{%
\begin{tabular}{lccccc|ccc}
\toprule
Method & HPSv3 $\uparrow$ & Vision. $\uparrow$ & Instruct. $\uparrow$ & MQ $\uparrow$ & Dynamic. $\uparrow$ & Total $\uparrow$ & Quality $\uparrow$ & Semantic $\uparrow$  \\
\midrule
LongLive & 8.44 & 14.31 & 62.04 & 18.84 & \textbf{70} & 83.91 & 84.71 & 80.70 \\
\rowcolor{LightBlue}
\hspace{1em}+ Ours &
\textbf{9.11}$_{\textcolor{ForestGreen}{+.67}}$ &
\textbf{14.93}$_{\textcolor{ForestGreen}{+.62}}$ &
\textbf{66.02}$_{\textcolor{ForestGreen}{+3.98}}$ & 
\textbf{19.37}$_{\textcolor{ForestGreen}{+.53}}$ &
64$_{\textcolor{red}{-6}}$ &
\textbf{84.51}$_{\textcolor{ForestGreen}{+.60}}$ &
\textbf{85.28}$_{\textcolor{ForestGreen}{+.57}}$ &
\textbf{81.44}$_{\textcolor{ForestGreen}{+.74}}$ \\
\bottomrule
\end{tabular}
}
\end{table}

\begin{figure}[t]
    \centering
    \includegraphics[width=1\linewidth]{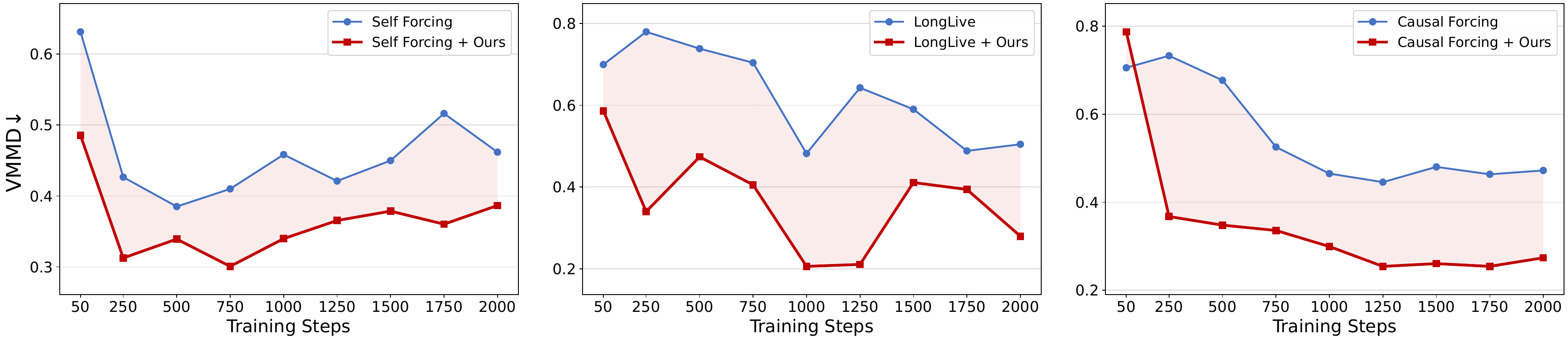}
    \caption{V-JEPA2 Maximum Mean Discrepancy (VMMD) over training steps on the evaluation set. Incorporating Mask Forcing leads to faster convergence across various baselines.}
    \label{fig:vjepa_mmd}
\end{figure}

\begin{table}[t]
\centering
\begin{minipage}[t]{0.26\columnwidth}
\centering
\caption{Mask ratio $\alpha$ ablation.}
\label{tab:alpha}
\resizebox{\linewidth}{!}{%
\begin{tabular}{lcc}
\toprule
$\alpha$ & HPSv3 $\uparrow$ & Dynamic. $\uparrow$ \\
\midrule
0.1 & 10.00 & 47 \\
\rowcolor{LightBlue}
0.2 & 9.84 & 82 \\
0.3 & 9.82 & 65 \\
0.4 & 10.15 & 57 \\
0.5 & 10.17 & 44 \\
\bottomrule
\end{tabular}
}
\end{minipage}
\hfill
\begin{minipage}[t]{0.26\columnwidth}
\centering
\caption{Timestep window $\Delta$ ablation.}
\label{tab:gap}
\resizebox{\linewidth}{!}{%
\begin{tabular}{lcc}
\toprule
$\Delta$ & HPSv3 $\uparrow$ & Dynamic. $\uparrow$ \\
\midrule
50 & 9.16 & 92 \\
150 & 9.82 & 70 \\
\rowcolor{LightBlue}
250 & 9.84 & 82 \\
450 & 9.98 & 80 \\
600 & 9.90 & 67 \\
\bottomrule
\end{tabular}
}
\end{minipage}
\hfill
\begin{minipage}[t]{0.42\columnwidth}
\centering
\caption{Mask scheme ablation.}
\label{tab:mask_scheme}
\resizebox{\linewidth}{!}{%
\begin{tabular}{lcc}
\toprule
Mask scheme & HPSv3 $\uparrow$ & Dynamic. $\uparrow$ \\
\midrule
shared, per-rollout & 10.12 & 56 \\
shared, per-chunk & 9.84 & 81 \\
shared, per-step & 9.64 & 72 \\
per-frame, per-rollout & 10.03 & 53 \\
\rowcolor{LightBlue}
per-frame, per-chunk & 9.84 & 82 \\
per-frame, per-step & 9.93 & 74 \\
\bottomrule
\end{tabular}
}
\end{minipage}
\end{table}

\subsection{Comparisons with Baselines}

\noindent \textbf{Quantitative Comparisons.} As shown in Tab.~\ref{tab:main}, Mask Forcing improves all baseline methods on both visual quality (HPSv3 \& Vision.) and semantic score (Instruct.) on the 100-prompt set under both chunk-wise and frame-wise settings. Especially for visual quality, the consistent improvements in visual quality align with our motivation of perturbing student rollout trajectories to alleviate artifacts associated with mode collapse. For motion evaluation, incorporating Mask Forcing produces higher motion quality (MQ) and dynamic degree (Dynamic.). Though the dynamic degree score for LongLive is lower, this can be attributed to generic VBench rewarding drift-induced optical flow~\citep{minar2025steady}. For the VBench benchmark, including Mask Forcing surpasses all baseline methods across the VBench metrics in both settings. Additionally, we evaluate Mask Forcing on single-prompt long video generation with LongLive, integrating it into both the initialization and streaming long-video tuning stages. As reported in Tab.~\ref{tab:long}, our method can also effectively improve visual quality in the long video generation setting.

\noindent \textbf{Qualitative Comparisons.} Fig.~\ref{fig:quali_1} presents the qualitative comparisons of our method and the baselines on the 100-prompt set and MovieGen. The baseline methods commonly exhibit limited visual quality, with over-saturation and over-smoothing artifacts. Specifically, Self Forcing produces the woman's face with unnaturally high-contrast coloration and the creature with a visually flat appearance, while LongLive generates scenes lacking fine-grained details. The rock-man and kitchen examples produced by Causal Forcing also exhibit severe over-saturation. Incorporating Mask Forcing effectively alleviates these issues, greatly enhancing the visual quality and realism of the generated videos. We additionally present the qualitative results for long video generation in Fig.~\ref{fig:quali_long}. Incorporating Mask Forcing produces higher visual quality with more high-frequency details, such as the fluffy fur and candle in case 1 and the fine-grained stone and sand textures in case 2.

\subsection{Ablation Studies}\label{sec:ablation}

We conduct comprehensive ablation studies on the 100-prompt set with Self Forcing as the baseline to validate the design choices in Mask Forcing. We jointly report HPSv3 and Dynamic Degree to measure visual quality and motion dynamics during analysis since models trained with DMD may suffer from reduced motion dynamics as training progresses. We further compare convergence speed across different baselines with and without Mask Forcing.

\noindent \textbf{Mask Ratio $\alpha$.} We compare different mask ratio values for the dual-noise masking rollout in Tab.~\ref{tab:alpha}. With a small ratio of 0.1, the masking trajectories only induce similar rollouts and provide limited exploration beyond the modes already covered by the student. Therefore, it achieves a high HPSv3 score of 10.00 but exhibits weak motion dynamics, with a Dynamic Degree of 47. With large ratios of 0.4-0.5, a substantial proportion of the input consists of cleaner states derived from the current student prediction. The resulting strong denoising cues yield the highest HPSv3 scores of 10.15 and 10.17. However, the Dynamic Degree decreases to 57 and 44, suggesting that excessive reliance on cleaner predictions may restrict the model’s ability to revise or diversify motion patterns during subsequent denoising. A moderate mask ratio of 0.2 introduces sufficient perturbation to diversify the student rollout trajectories while avoiding excessive reliance on cleaner states derived from the current student prediction. It achieves a competitive HPSv3 score of 9.84 and the highest Dynamic Degree of 82, providing a balance between visual quality and motion dynamics.

\noindent \textbf{Timestep Window $\Delta$.} We ablate the timestep window $\Delta$ used to sample the timestep corresponding to a lower noise level in Tab.~\ref{tab:gap}. This window controls the maximum difference between original and cleaner noise levels. With a small timestep window of $\Delta=50$, the cleaner noise level remains close to the original, providing only mild rollout perturbations and limited denoising cues. Therefore, this setting achieves a high Dynamic Degree of $92$ but a low HPSv3 score of $9.16$. Increasing $\Delta$ strengthens the rollout perturbation and provides cleaner tokens with more informative denoising cues. Consequently, increasing $\Delta$ to 150, 250, and 450 improves the HPSv3 score to 9.82, 9.84, 9.98, respectively. However, an excessively large window of $\Delta = 600$ may cause cleaner predictions to dominate subsequent denoising, limiting motion diversity. Therefore, we adopt $\Delta = 250$, which achieves a competitive HPSv3 score of 9.84 while maintaining a high Dynamic Degree of 82.

\noindent \textbf{Mask Scheme.} We ablate the influence of different mask schemes for chunk-wise generation in Tab.~\ref{tab:mask_scheme}. We evaluate the schemes varied along spatial and temporal axes: across the frames within each chunk and across chunks during the rollout. Spatially, ``shared" applies the same mask to all frames within a chunk and ``per-frame" independently samples a mask for each frame. Temporally, ``per-rollout" keeps the mask fixed throughout the entire rollout, “per-chunk” resamples the mask for each chunk while keeping it fixed across the denoising steps within the chunk, and “per-step” resamples the mask at every denoising step. These spatial and temporal mask choices determine how the mask-induced perturbations vary throughout the rollout and consequently affect the resulting student rollout distributions. An effective scheme should sufficiently diversify the student distribution to improve its coverage of the teacher modes without introducing excessive variation that deviates from the teacher distribution. As shown in Tab.~\ref{tab:mask_scheme}, applying ``per-frame" in the spatial axis and ``per-chunk" in the temporal axis achieves both high visual quality and motion dynamics, with a HPSv3 score of 9.84 and Dynamic Degree of 82.

\noindent \textbf{Convergence Speed.} We compare the training convergence of different baselines with and without Mask Forcing in Figs.~\ref{fig:teaser},~\ref{fig:vjepa_mmd} and~\ref{fig:longlive_hps_mmd}. We use HPSv3 to evaluate the visual quality trend throughout the training. To assess the distributional alignment between student- and teacher-generated videos, we additionally report Maximum Mean Discrepancy (MMD)~\citep{jayasumana2024rethinking} in the CLIP~\citep{radford2021learning} and V-JEPA2~\citep{assran2025v} feature spaces, denoted as CMMD and VMMD, respectively. Specifically, we first generate a fixed reference set with the real-score teacher on the 100-prompt set. We then generate videos for each step using the causal student on the same prompts and compute CMMD and VMMD against the teacher reference set. As shown in Figs.~\ref{fig:teaser},~\ref{fig:vjepa_mmd} and~\ref{fig:longlive_hps_mmd}, Mask Forcing accelerates convergence for all baselines in terms of both visual quality and distributional alignment.

\subsection{Human Evaluation}
\begin{figure}[t]
    \centering
    \includegraphics[width=1\linewidth]{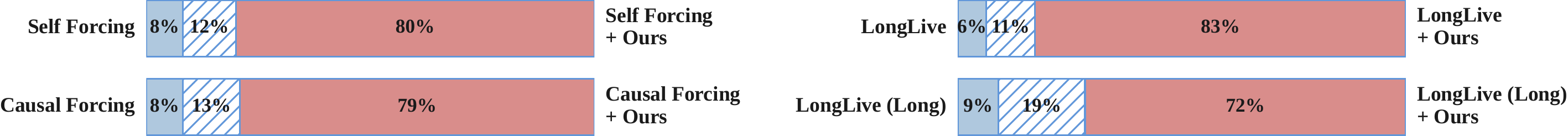}
    \caption{Pairwise human preferences between baseline models with and without Mask Forcing. Blue and red denote preferences for models without and with Mask Forcing, respectively, while hatched regions indicate ties.}
    \label{fig:user_1}
    \vspace{-0.5em}
\end{figure}

We conduct pairwise human evaluations between anonymized outputs from Mask Forcing (+Ours) against different baselines for short and long video generation. We recruit 24 users to conduct this human evaluation study, and each user is required to select the video with higher visual quality and greater realism from each pair of videos. As shown in Fig.~\ref{fig:user_1}, incorporating Mask Forcing receives 80\%, 79\%, and 83\% of the preference votes over the baseline methods Self Forcing, Causal Forcing, and LongLive, respectively. Moreover, for long video generation, Mask Forcing receives 72\% of the preference votes over the LongLive baseline, further demonstrating the effectiveness of our method.

\section{Conclusion}

In this paper, we introduce Mask Forcing, a Dual-Noise Masking Rollout strategy for AR video diffusion distillation that alleviates the mode collapse induced by the mode-seeking reverse KL objective in DMD. Mask Forcing injects lower-noise signals into the noisy rollout inputs via random masks varying along spatial and temporal axes. These perturbations promote broader exploration during self-rollout, diversifying student trajectories and encouraging the student distribution to cover more regions of the teacher distribution and thereby allowing DMD to provide learning signals beyond the modes already covered by the student. Meanwhile, cleaner tokens act as denoising guidance for noisier tokens, improving intermediate rollout predictions and mitigating error accumulation during the self-rollout process. Extensive experiments demonstrate the effectiveness of our method across multiple baselines, with improved visual quality and convergence speed. We further investigate the effects of different masking mechanisms on the visual quality and motion dynamics balance on the generated videos. Overall, Mask Forcing provides a simple and effective strategy to improve AR video diffusion distillation without incorporating real video data or post-training.

\clearpage
\bibliography{main}
\bibliographystyle{main}

\section{Appendix}

\subsection{Theoretical Justification}\label{app:theory}

In this section, we characterize the distribution induced by Mask Forcing and derive an exact decomposition of its reverse-KL objective. Fix a text condition $c$ and a DMD score-noising timestep $\tau>0$. Let
\begin{equation}
    p_\tau(x):=p_{\mathrm{real},\tau}(x\mid c)
\end{equation}
denote the teacher noisy marginal. Let
\begin{equation}
    V:=\{(M_r,t_{k,r})\}_{r=1}^{R},
    \qquad V\sim\pi,
\end{equation}
collect the masks and cleaner timesteps sampled during one Mask Forcing rollout, where $\pi$ is the distribution induced by the corresponding samplers.

Given a masking trajectory $V=v$, define the conditional student distribution as
\begin{equation}
    q^v_{\theta,\tau}(x)
    :=p_\theta(X_\tau=x\mid V=v,c).
\end{equation}
Marginalizing over $V$ gives
\begin{equation}
    \bar q_{\theta,\tau}(x)
    :=\mathbb E_{V\sim\pi}
    \left[q^V_{\theta,\tau}(x)\right].
    \label{eq:mask-mixture}
\end{equation}
Under exact score estimation, the real and fake scores at noise level $\tau$ are
\begin{equation}
    s_{\mathrm{real}}(x,\tau,c)
    =\nabla_x\log p_\tau(x),
    \qquad
    s_{\mathrm{fake}}(x,\tau,c)
    =\nabla_x\log\bar q_{\theta,\tau}(x).
\end{equation}
Since the fake-score model is not conditioned on $V$, it estimates the score of the marginal distribution $\bar q_{\theta,\tau}$ rather than a component score $\nabla_x\log q^v_{\theta,\tau}$.

The DMD score-difference update corresponds to the gradient
\begin{equation}
\begin{aligned}
    \nabla_\theta
    D_{\mathrm{KL}}
    \left(\bar q_{\theta,\tau}\|p_\tau\right)
    =\mathbb E
    \left[
        \left(
            s_{\mathrm{fake}}(X_\tau,\tau,c)
            -s_{\mathrm{real}}(X_\tau,\tau,c)
        \right)^{\!\top}
        \nabla_\theta X_\tau
    \right].
\end{aligned}
\label{eq:dmd-kl-gradient}
\end{equation}
We therefore analyze the reverse-KL objective
\begin{equation}
    D_{\mathrm{KL}}
    \left(\bar q_{\theta,\tau}\|p_\tau\right).
    \label{eq:masked-dmd-kl}
\end{equation}
The exact-score assumption is used only for this identification. In practice, both score networks approximate the corresponding marginal scores.

\paragraph{Proposition (KL decomposition for the masked-rollout distribution).}
\label{prop:mixture-decomposition}
Suppose that the KL divergences below are finite. Then
\begin{equation}
    D_{\mathrm{KL}}
    \left(\bar q_{\theta,\tau}\|p_\tau\right)
    =
    \mathbb E_V
    \left[
        D_{\mathrm{KL}}
        \left(q^V_{\theta,\tau}\|p_\tau\right)
    \right]
    -I_\theta(V;X_\tau\mid c).
\label{eq:mixture-decomposition}
\end{equation}

\paragraph{Proof.}
By the definition of KL divergence, the average conditional KL can be expanded by multiplying and dividing the density ratio by $\bar q_{\theta,\tau}$:
\begin{align}
&\mathbb E_V
D_{\mathrm{KL}}
\left(q^V_{\theta,\tau}\|p_\tau\right)
\notag\\
&=\int \pi(v)
\int q^v_{\theta,\tau}(x)
\log
\frac{q^v_{\theta,\tau}(x)}
     {p_\tau(x)}
\,\mathrm{d}x\,\mathrm{d}v
\notag\\
&=\mathbb E_{V,X_\tau\mid c}
\left[
    \log
    \frac{q^V_{\theta,\tau}(X_\tau)}
         {p_\tau(X_\tau)}
\right]
\notag\\
&=\mathbb E_{V,X_\tau\mid c}
\left[
    \log
    \frac{q^V_{\theta,\tau}(X_\tau)}
         {\bar q_{\theta,\tau}(X_\tau)}
    +
    \log
    \frac{\bar q_{\theta,\tau}(X_\tau)}
         {p_\tau(X_\tau)}
\right]
\notag\\
&=\mathbb E_{V,X_\tau\mid c}
\left[
    \log
    \frac{q^V_{\theta,\tau}(X_\tau)}
         {\bar q_{\theta,\tau}(X_\tau)}
\right]
+
\mathbb E_{V,X_\tau\mid c}
\left[
    \log
    \frac{\bar q_{\theta,\tau}(X_\tau)}
         {p_\tau(X_\tau)}
\right]
\notag\\
&=\underbrace{
    \mathbb E_V
    D_{\mathrm{KL}}
    \left(q^V_{\theta,\tau}\|\bar q_{\theta,\tau}\right)
}_{I_\theta(V;X_\tau\mid c)}
+
D_{\mathrm{KL}}
\left(\bar q_{\theta,\tau}\|p_\tau\right).
\label{eq:conditional-kl-decomposition}
\end{align}
The first term is the conditional mutual information. By definition,
\begin{align}
I_\theta(V;X_\tau\mid c)
&=\mathbb E_{V,X_\tau\mid c}
\left[
    \log
    \frac{p_\theta(V,X_\tau\mid c)}
         {p(V\mid c)p_\theta(X_\tau\mid c)}
\right]
\notag\\
&=\int \pi(v)q^v_{\theta,\tau}(x)
\log
\frac{\pi(v)q^v_{\theta,\tau}(x)}
     {\pi(v)\bar q_{\theta,\tau}(x)}
\,\mathrm{d}x\,\mathrm{d}v
\notag\\
&=\int \pi(v)
\left[
    \int q^v_{\theta,\tau}(x)
    \log
    \frac{q^v_{\theta,\tau}(x)}
         {\bar q_{\theta,\tau}(x)}
    \,\mathrm{d}x
\right]\mathrm{d}v
\notag\\
&=\int \pi(v)
D_{\mathrm{KL}}
\left(q^v_{\theta,\tau}\|\bar q_{\theta,\tau}\right)
\,\mathrm{d}v
\notag\\
&=\mathbb E_V
D_{\mathrm{KL}}
\left(q^V_{\theta,\tau}\|\bar q_{\theta,\tau}\right).
\label{eq:mi-as-average-kl}
\end{align}
For the second term, we first expand the joint expectation and then marginalize over $V$:
\begin{align}
&\mathbb E_{V,X_\tau\mid c}
\left[
    \log\frac{\bar q_{\theta,\tau}(X_\tau)}
                   {p_\tau(X_\tau)}
\right]
\notag\\
&=\int \pi(v)
\left[
    \int q^v_{\theta,\tau}(x)
    \log\frac{\bar q_{\theta,\tau}(x)}
                   {p_\tau(x)}
    \,\mathrm{d}x
\right]\mathrm{d}v
\notag\\
&=\int
\left[
    \int \pi(v)q^v_{\theta,\tau}(x)
    \,\mathrm{d}v
\right]
\log\frac{\bar q_{\theta,\tau}(x)}
         {p_\tau(x)}
\,\mathrm{d}x
\notag\\
&=\int \bar q_{\theta,\tau}(x)
\log\frac{\bar q_{\theta,\tau}(x)}
         {p_\tau(x)}
\,\mathrm{d}x
\notag\\
&=D_{\mathrm{KL}}
\left(\bar q_{\theta,\tau}\|p_\tau\right).
\label{eq:marginal-kl}
\end{align}
Combining the two terms and rearranging proves \eqref{eq:mixture-decomposition}.

The first term in \eqref{eq:mixture-decomposition} is the average reverse KL of the conditional rollout distributions. The mutual information measures their dependence on the masking trajectory. It also follows from \eqref{eq:mixture-decomposition} that
\begin{equation}
    D_{\mathrm{KL}}
    \left(\bar q_{\theta,\tau}\|p_\tau\right)
    \le
    \mathbb E_V
    D_{\mathrm{KL}}
    \left(q^V_{\theta,\tau}\|p_\tau\right).
\end{equation}
Since mutual information is nonnegative, equality holds when $V$ and $X_\tau$ are conditionally independent given $c$, meaning that the masking trajectory does not affect the conditional output distribution. If different masking trajectories induce distinguishable conditional output distributions, then $I_\theta(V;X_\tau\mid c)>0$, and the inequality is strict.

\paragraph{Implication for mode coverage.}
Under limited model capacity, reverse-KL optimization may concentrate a conditional rollout distribution on a high-density region of the teacher distribution. If different masking trajectories induce distinct conditional distributions, then $I_\theta(V;X_\tau\mid c)>0$, and \eqref{eq:mixture-decomposition} shows that their marginal mixture has strictly lower reverse KL than their average conditional reverse KL. This mixture can cover multiple teacher-supported regions without requiring each conditional distribution to place probability mass between them. The result therefore gives a distributional mechanism by which randomized masking can alleviate mode-seeking behavior.

\subsection{Comparison with Joint Distillation}

Recent methods seek to improve distillation performance by combining complementary objectives that balance mode-seeking and mode-covering behaviors, such as rCM and DistillAlign. We compare with Causal-rCM~\citep{zheng2026causal} and DistillAlign in terms of visual quality and motion dynamics on the 100-prompt set benchmark. While rCM jointly optimizes continuous-time consistency distillation and score-based distillation in bidirectional models, Causal-rCM decouples these objectives into two stages. It first performs consistency distillation under teacher forcing to obtain causal initialization with broad mode coverage, and then applies DMD under self-forcing. Since DistillAlign is built upon Causal Forcing and Causal-rCM pairs its two distillation objectives with two causal training paradigms, we compare Mask Forcing on the Causal Forcing baseline with them. For DistillAlign, we use its final joint-distilled generator with the Wan2.1-T2V-14B teacher. For Causal-rCM, we follow its 4-step inference setting. 

\begin{figure}[t]
    \centering
    \includegraphics[width=\linewidth]{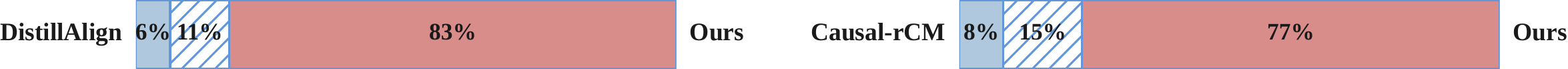}
    \caption{Pairwise human preference between joint distillation methods (DistillAlign and Causal-rCM) and Mask Forcing.}
    \label{fig:user_2}
\end{figure}

\begin{figure}[t]
    \centering
    \includegraphics[width=1\linewidth]{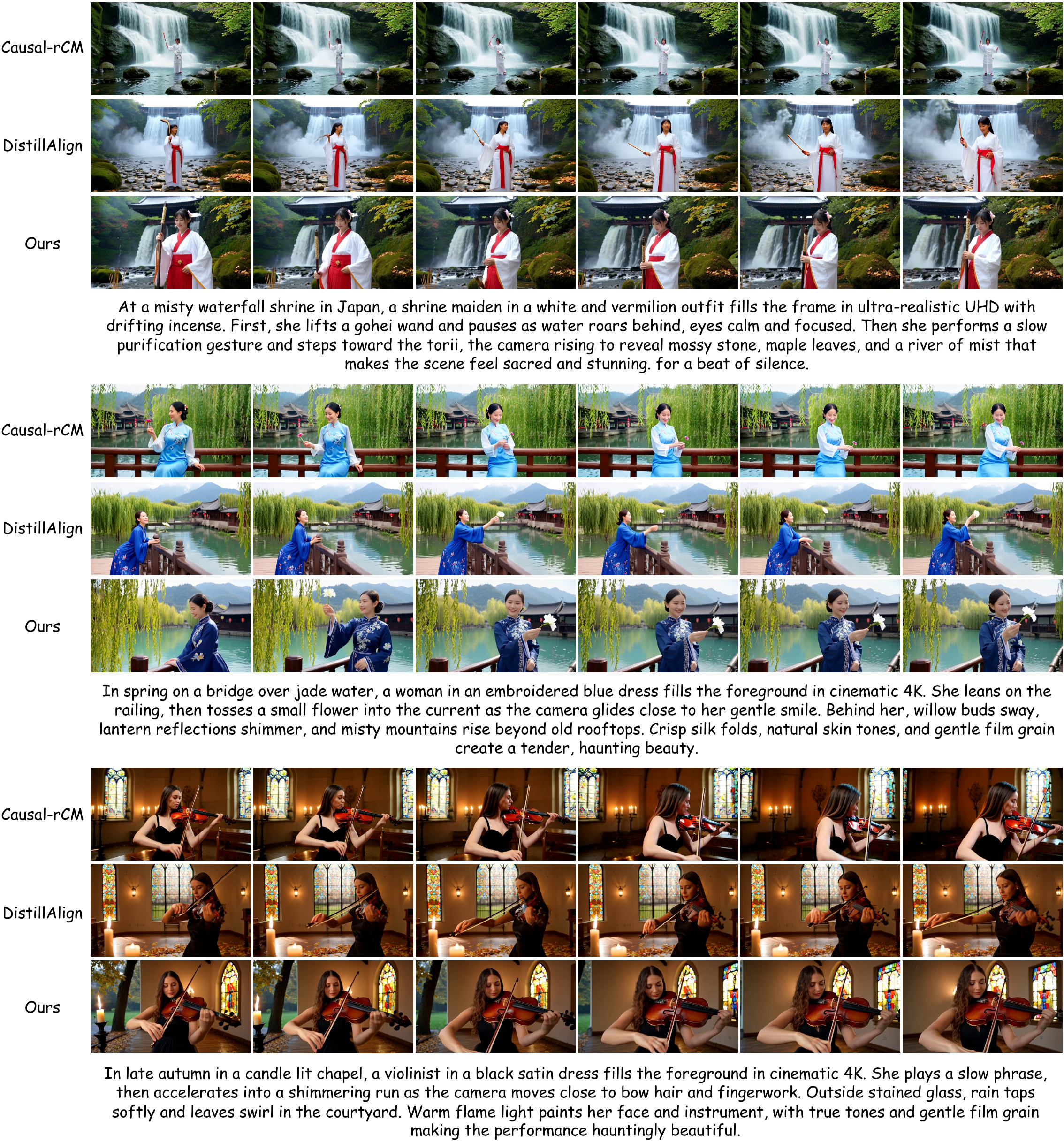}
    \caption{Qualitative comparison of our method with Causal-rCM and DistillAlign. Our method produces videos with better visual quality and motion dynamics.}
    \label{fig:comp_joint_distill}
\end{figure}

As shown in Tab.~\ref{tab:distill}, our method outperforms both methods in visual quality, motion dynamics, as well as semantic alignment. Qualitative comparison in Fig.~\ref{fig:comp_joint_distill} is consistent with the quantitative results, showing that our method generates videos with richer details and greater realism. We further conduct pairwise human preference against joint distillation methods DistillAlign and Causal-rCM. As shown in Fig.~\ref{fig:user_2}, Mask Forcing receives more preference votes, with 83\% and 77\% over DistillAlign and Causal-rCM, respectively. These results demonstrate that Mask Forcing yields videos with higher visual quality and realism from the users' perspective.

\begin{table}[t]
\centering
\caption{Quantitative comparison of Mask Forcing (Ours) with Causal-rCM and DistillAlign on the 100-prompt set benchmark.}
\label{tab:distill}
\resizebox{1\linewidth}{!}{%
\begin{tabular}{lccccc}
\toprule
Method & HPSv3 $\uparrow$ & Vision. $\uparrow$ & Instruct. $\uparrow$ & MQ $\uparrow$ & Dynamic. $\uparrow$ \\
\midrule
DistillAlign~\citep{li2026distillalign} & 9.29 & 9.46 & 37.69 & 14.36 & 76  \\
Causal-rCM~\citep{zheng2026causal}  & 9.61 & 9.43 & 38.50 & 14.34 & 64  \\
Ours & \textbf{10.17} & \textbf{11.58} & \textbf{46.30} & \textbf{21.54} & \textbf{82} \\
\bottomrule
\end{tabular}
}
\end{table}

\subsection{Additional Evaluation on LongLive}

\begin{figure}[t]
    \centering
    \includegraphics[width=0.9\linewidth]{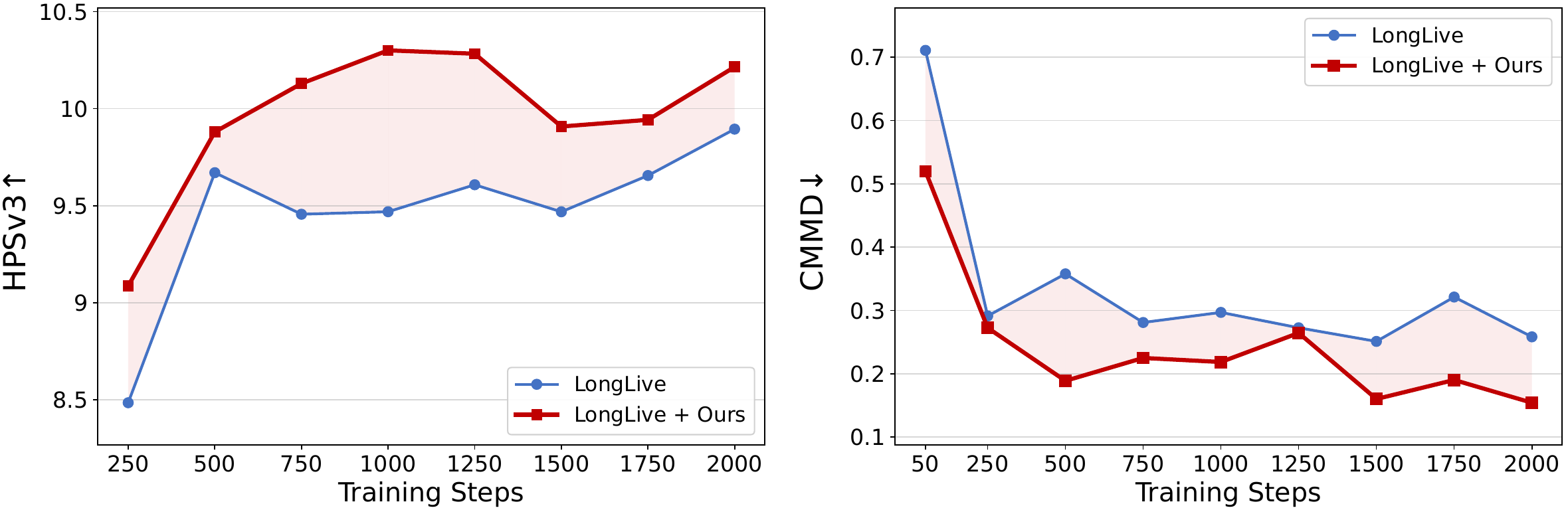}
    \caption{HPSv3 and CMMD over training steps on the evaluation set for LongLive and LongLive (+Ours).}
    \label{fig:longlive_hps_mmd}
\end{figure}

We compare the convergence of LongLive with and without Mask Forcing in Fig.~\ref{fig:longlive_hps_mmd}. Consistent with the results on Self Forcing and Causal Forcing, Mask Forcing achieves higher HPSv3 and lower CMMD in fewer training steps, demonstrating faster convergence.

Moreover, we evaluate LongLive and Mask Forcing on MovieGen over consecutive 6-second intervals from 0 to 30 seconds. For each interval, we average the CLIP score over all frames and the HPSv3 score over 12 uniformly sampled frames. As reported in Tab.~\ref{tab:inter_clip_hps}, both methods exhibit a certain decline in CLIP and HPSv3 scores as the video generation progresses, likely due to error accumulation during autoregressive self-rollout. However, incorporating Mask Forcing achieves higher CLIP and HPSv3 scores than LongLive across all intervals, demonstrating consistent improvements in semantic alignment and visual quality throughout long video generation.

\begin{wraptable}{r}{0.45\linewidth}
\centering
\caption{Video diversity across baselines.}
\label{tab:div}
\resizebox{\linewidth}{!}{%
\begin{tabular}{lcc}
\toprule
Method & CLIP Div.$\uparrow$ & DINO Div.$\uparrow$ \\
\midrule
Self Forcing 
& 0.0773 &  \textbf{0.1704} \\
\rowcolor{LightBlue}
\hspace{1em}+ Ours 
&  \textbf{0.0814} & 0.1645 \\
LongLive 
& 0.0784  & 0.1537  \\
\rowcolor{LightBlue}
\hspace{1em}+ Ours 
&  \textbf{0.0819} & \textbf{0.1750} \\
Causal Forcing 
& 0.0675 & 0.1371 \\
\rowcolor{LightBlue}
\hspace{1em}+ Ours 
&  \textbf{0.0720} & \textbf{0.1577} \\
\bottomrule
\end{tabular}
}
\end{wraptable}

\subsection{Diversity Evaluation}
We further evaluate the video diversity across different baselines under the chunk-wise setting using CLIP-ViT-Large~\citep{radford2021learning} and DINOv3-ViT-Large~\citep{simeoni2025dinov3} following DP-DMD~\citep{wu2026diversitypreserved}. We use the 100-prompt set as the evaluation set and generate 8 videos for each prompt with random seeds 0-7, generating 800 videos for each baseline for evaluation in total.

As shown in Tab.~\ref{tab:div}, Mask Forcing improves CLIP diversity across all three baselines, indicating consistently greater semantic variation. It also substantially improves DINOv3 diversity for Causal Forcing and LongLive, suggesting increased structural variation. For Self Forcing, DINOv3 diversity decreases slightly despite the improvement in CLIP diversity. Together with the consistent gains in visual quality, semantic alignment, distributional alignment, and human preference, these results suggest that Mask Forcing jointly improves generation quality and diversity.

\subsection{Interactive Video World Model Application}

\begin{figure}[t]
    \centering
    \includegraphics[width=1\linewidth]{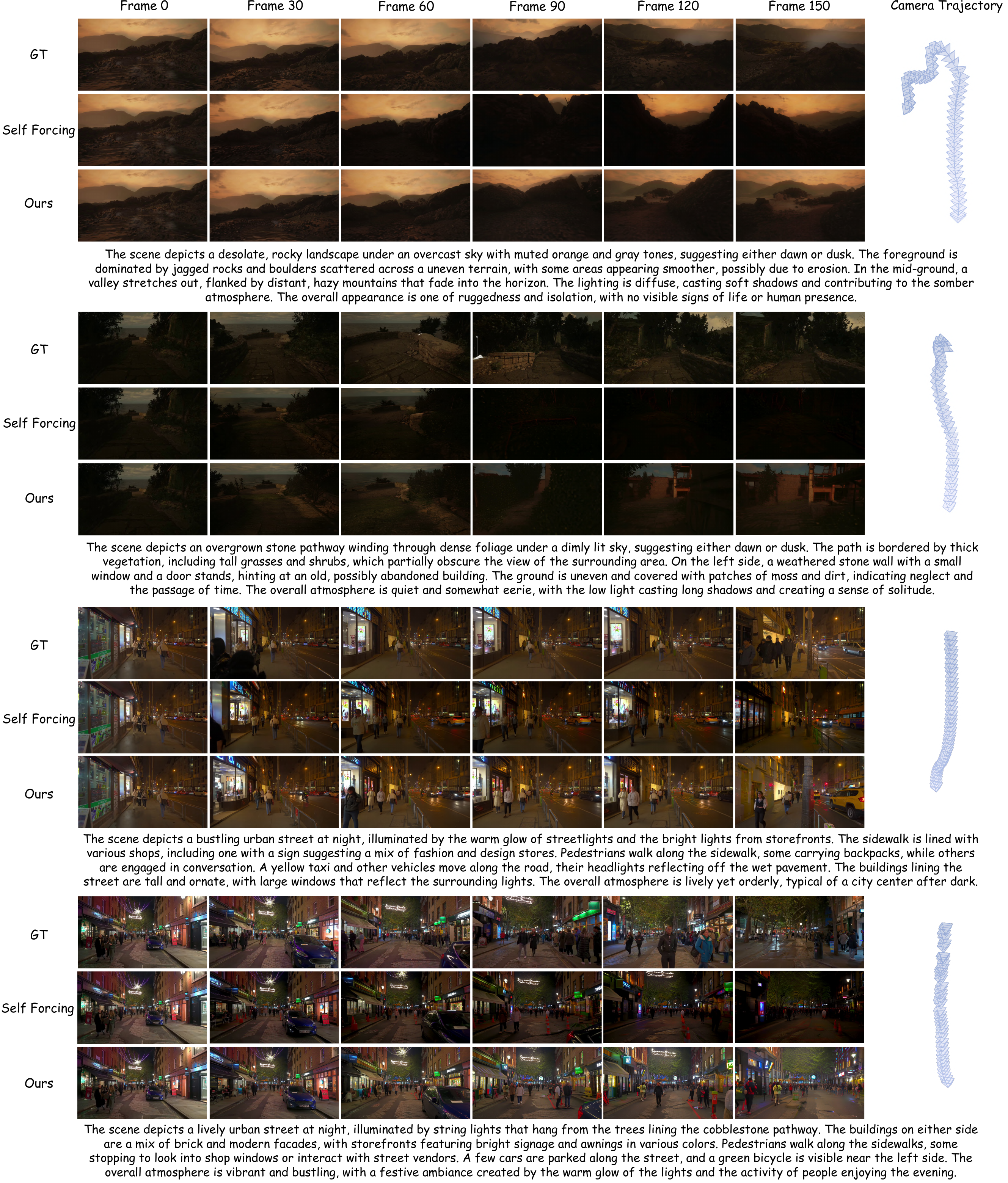}
    \caption{Qualitative comparison of Self Forcing and Mask Forcing (Ours) on Sekai-Game and Sekai-Walking. As the rollout progresses, Self Forcing videos darken and lose fine details. Mask Forcing mitigates this degradation, preserving rich details throughout generation.}
    \label{fig:wm}
\end{figure}

We further explore Mask Forcing on autoregressive video generation with camera control. Current interactive world models extend video generation by conditioning future observations on user inputs~\citep{team2026dreamx, team2026advancing, hyworld2025, wang2026matrix, zhao2026minwm}, such as actions, camera motions, or instructions. We evaluate Mask Forcing in a camera-controlled image-to-video setting, where an input image serves as the initial frame, and the model generates subsequent frames following a specified camera trajectory. We utilize Wan2.2-5B-TI2V as the base model and train the bidirectional model with open-source data, such as SpatialVID~\citep{wang2026spatialvid}, OmniWorld~\citep{zhou2025omniworld}, RealCam-Vid~\citep{zheng2025realcam}, DL3DV~\citep{ling2024dl3dv}, Sekai~\citep{li2026sekai}, and MiraData~\citep{ju2024miradata}. Then we train an AR model using teacher forcing and adopt consistency distillation to obtain a few-step generator. Both bidirectional and autoregressive models are trained on 5s videos following SolarWM~\citep{huang2026solarwm}.  We compare Self Forcing and Mask Forcing for further causal student distillation to align with the teacher's distribution with DMD. Fig.~\ref{fig:wm} presents qualitative comparisons on the Sekai-Game and Sekai-Walking test set. Videos generated by Self Forcing become darker and lose fine-grained details as the rollout proceeds. This issue can be attributed to the mode-seeking behavior of the reverse KL objective in DMD and error accumulation during self-rollout. Mask Forcing mitigates this degradation, producing scenes with richer details throughout the rollout. We view this experiment as an initial evaluation of our method in the camera-controlled image-to-video setting. To enable long and precise interactive video generation, future work includes training with long sequences and rollouts~\citep{yang2025longlive} and the context update strategy for temporal extrapolation~\citep{zhuang2026selfgradientforcingnative}.

\begin{table}[t]
\centering
\caption{Quantitative evaluation on long video generation on MovieGen. We evaluate the CLIP Score and HPSv3 across 6-second intervals (0--30s).}
\label{tab:inter_clip_hps}
\resizebox{1\linewidth}{!}{%
\begin{tabular}{lccccc!{\vrule width 0.5pt}ccccc}
\toprule
\multirow{2}{*}{Method} 
& \multicolumn{5}{c}{CLIP$\uparrow$} 
& \multicolumn{5}{c}{HPSv3$\uparrow$} \\
\cmidrule(lr){2-6} \cmidrule(lr){7-11}
& 0-6 & 6-12 & 12-18 & 18-24 & 24-30
& 0-6 & 6-12 & 12-18 & 18-24 & 24-30 \\
\midrule
LongLive 
& 33.70 & 33.41 & 33.29 & 33.23 & 33.15 
& 8.81 & 8.37 & 8.23 & 7.89 & 7.79 \\
\hspace{1em}+ Ours 
& \textbf{34.12} & \textbf{33.81} & \textbf{33.59} & \textbf{33.63} & \textbf{33.38} 
& \textbf{9.65} & \textbf{9.44} & \textbf{9.05} & \textbf{8.87} & \textbf{8.68} \\
\bottomrule
\end{tabular}
}
\end{table}

\subsection{Algorithm}

We provide detailed pseudocode for training with Mask Forcing in Algorithm~\ref{alg:mask-forcing}. Mask Forcing requires no additional forward passes, external data, or post-training stages and can be seamlessly integrated into existing self-rollout training pipelines. Mask Forcing constructs a dual-noise input at each denoising step by mixing the original- and lower-noise states with a random mask. The perturbed input is then denoised at the original timestep, and the final clean prediction is cached as context for subsequent blocks.

\begin{algorithm}[t]
\caption{Mask Forcing Training}
\label{alg:mask-forcing}
\begin{algorithmic}[1]
\Require Denoising timesteps $\mathcal{T}=\{t_1,\dots,t_T\}$ with $t_T=1$ and clean endpoint $t_0=0$
\Require Text prompt $c$; number of autoregressive chunks $F$, latent frames per chunk $L$, spatial latent resolution $H\!\times\!W$
\Require AR diffusion model $G_\theta$ (returns KV embeddings via $G^{\mathrm{KV}}_\theta$)
\Require Mask ratio $\alpha$, timestep window size $\Delta$ in training timestep units, number of training timesteps $N_t$, normalized timestep floor $t_{\min}$
\Statex \textbf{Notation:}\ $\Psi(x,\varepsilon,t)=(1-t)x+t\varepsilon$
\State \textbf{loop}
\State \quad Initialize KV cache $\mathrm{KV} \gets [\,]$ and model output $X_\theta \gets [\,]$
\State \quad Sample exit index $s \sim \mathrm{Uniform}\{1,\dots,T\}$
\For{chunk $i = 1,\dots,F$}
    \State Sample initial noisy latent $x_i^{t_T} \sim \mathcal{N}(0, I)$
    \State Sample a per-frame binary mask $M^i\in\{0,1\}^{L\times H\times W}$ with ratio $\alpha$
    \If{$s=T$}
        \State Enable gradient computation
    \Else
        \State Disable gradient computation
    \EndIf
    \State $\hat{x}_{i,T}^0 \gets G_\theta\!\left(x_i^{t_T};\, c,\, t_T,\, \mathrm{KV}\right)$ \Comment{initial unperturbed prediction}
    \If{$s<T$}
        \For{denoising step $j = T-1,T-2,\dots,s$}
            \State $t'_k \sim \mathrm{Uniform}\!\left([\max(t_{\min},t_j-\Delta/N_t),\,t_j]\right)$
            \State Sample $\varepsilon_j \sim \mathcal{N}(0, I)$
            \State $x_i^{t'_k} \gets \Psi(\hat{x}_{i,j+1}^0,\varepsilon_j,t'_k)$, \quad $x_i^{t_j} \gets \Psi(\hat{x}_{i,j+1}^0,\varepsilon_j,t_j)$
            \State $x_i^{t_{\mathrm{mix}}} \gets M^i\odot x_i^{t'_k}+(1-M^i)\odot x_i^{t_j}$
            \If{$j=s$}
                \State Enable gradient computation
            \Else
                \State Disable gradient computation
            \EndIf
            \State $\hat{x}_{i,j}^0 \gets G_\theta\!\left(x_i^{t_{\mathrm{mix}}};\,c,\,t_j,\,\mathrm{KV}\right)$ \Comment{perturbed prediction at step $j$}
        \EndFor
    \EndIf
    \State Append $\hat{x}_{i,s}^0$ to $X_\theta$
    \State Disable gradient computation
    \State $\mathrm{kv}^i \gets G^{\mathrm{KV}}_\theta\!\left(\hat{x}_{i,s}^0;\,c,\,t_0,\,\mathrm{KV}\right)$ \Comment{cache the exit prediction}
    \State $\mathrm{KV}.\mathrm{append}(\mathrm{kv}^i)$
\EndFor
\State \quad Update $\theta$ via the distribution matching loss $\mathcal{L}_{\mathrm{DMD}}(X_\theta)$
\State \textbf{end loop}
\end{algorithmic}
\end{algorithm}

\subsection{More Qualitative Results}

\begin{figure}[t]
    \centering
    \includegraphics[width=1\linewidth]{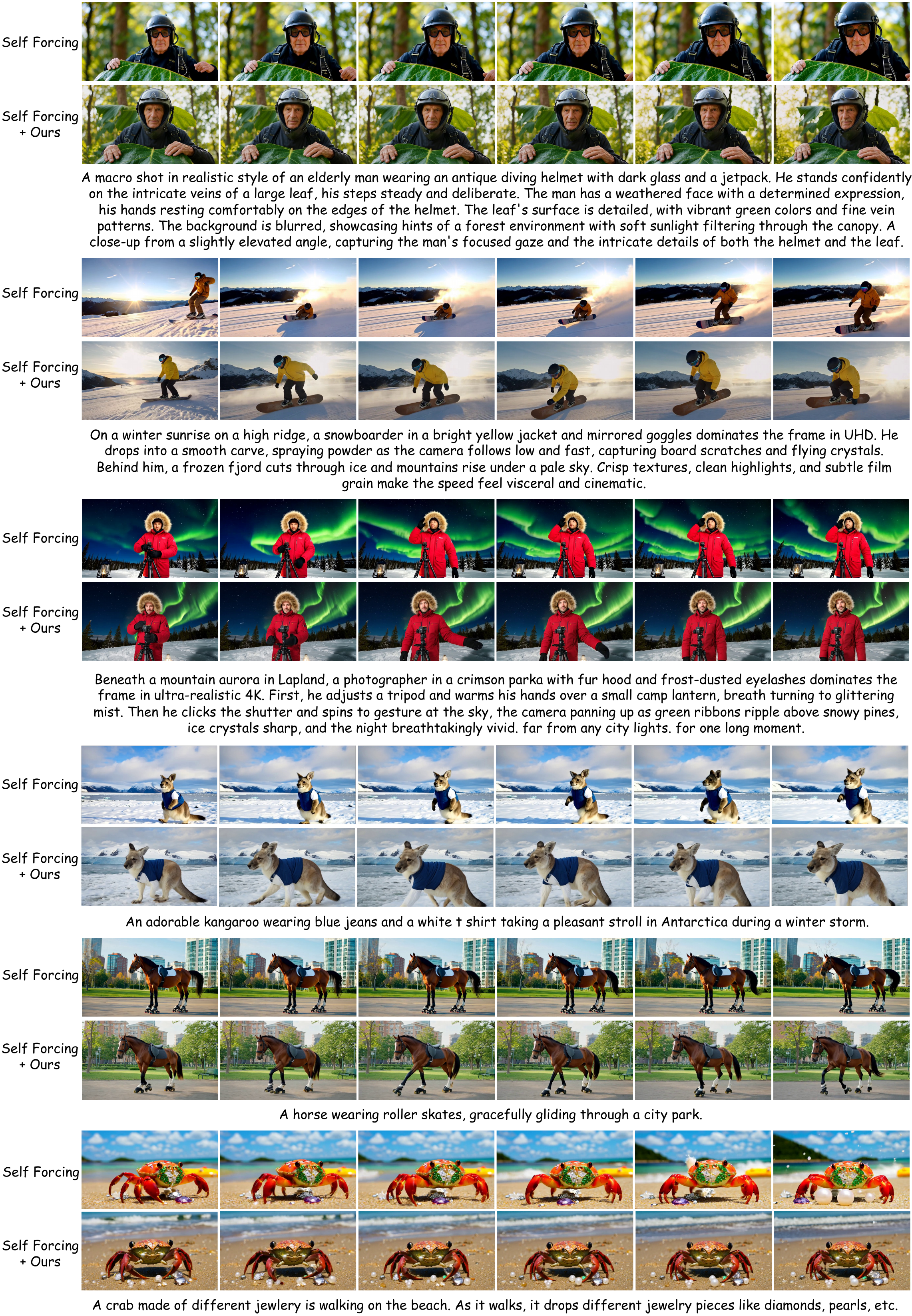}
    \caption{Additional qualitative comparisons between Self Forcing and Self Forcing (+Ours).}
    \label{fig:app_sf}
\end{figure}

\begin{figure}[t]
    \centering
    \includegraphics[width=1\linewidth]{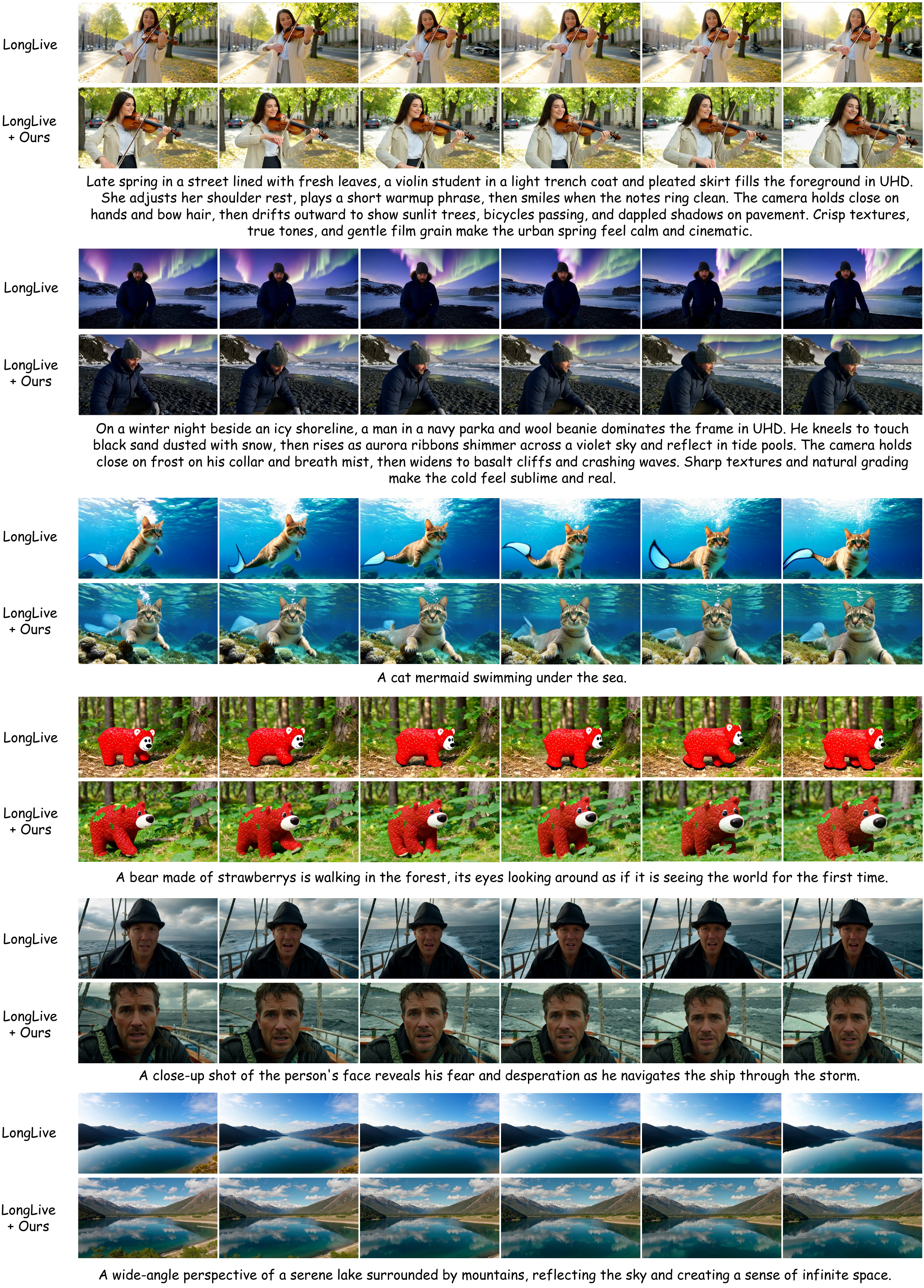}
    \caption{Additional qualitative comparisons between LongLive and LongLive (+Ours).}
    \label{fig:app_ll}
\end{figure}

\begin{figure}[t]
    \centering
    \includegraphics[width=1\linewidth]{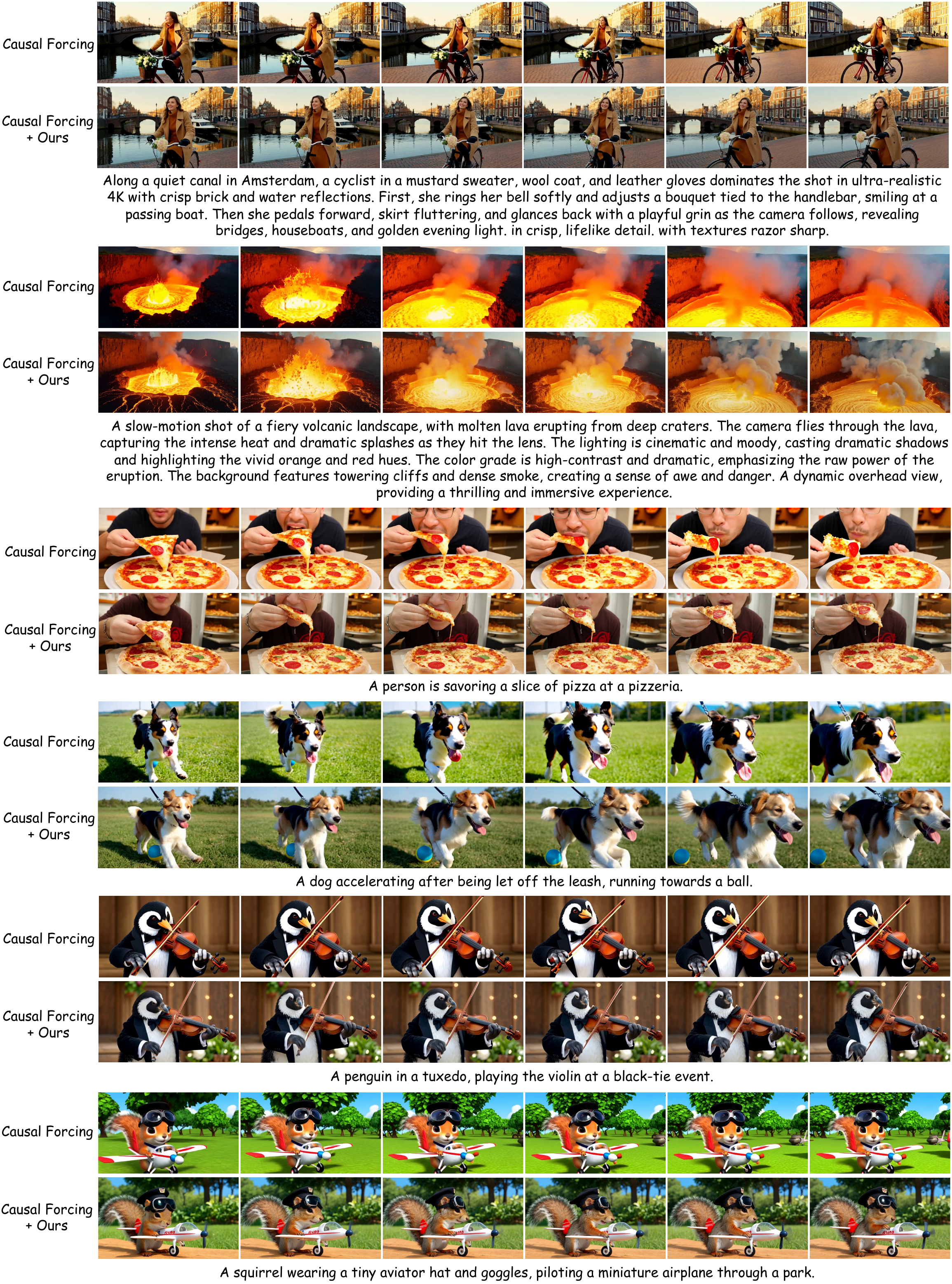}
    \caption{Additional qualitative comparisons between Causal Forcing and Causal Forcing (+Ours).}
    \label{fig:app_cf}
\end{figure}

We present additional qualitative comparisons on single-prompt short video generation on the 100-prompt set and MovieGen in Figs.~\ref{fig:app_sf}-\ref{fig:app_cf}. Qualitative comparisons across different methods show that Mask Forcing consistently improves the visual quality of the generated videos, enhancing them with greater realism and richer detail while preventing over-saturated and over-smoothed artifacts.

\end{document}